\documentclass{article}

\usepackage{microtype}
\usepackage{graphicx}
\usepackage{subcaption}
\usepackage{booktabs} 
\usepackage{hyperref}
\usepackage{algorithmic}
\usepackage{url}
\usepackage{amsmath,amssymb}
\usepackage{xcolor}
\usepackage{hyperref}
\usepackage{nameref}
\definecolor{HeaderBlue}{RGB}{230,243,255}
\definecolor{BandGray}{gray}{0.95}
\definecolor{BandBlue}{RGB}{245,250,255}
\definecolor{GroupGray}{gray}{0.93}
\definecolor{OursGreen}{RGB}{235,247,236}
\usepackage{colortbl}

\usepackage{amsmath}
\usepackage[utf8]{inputenc} 
\usepackage[T1]{fontenc}    
\usepackage{hyperref}       
\usepackage{url}            
\usepackage{booktabs}       
\usepackage{amsfonts}       
\usepackage{nicefrac}       
\usepackage{microtype}      
\usepackage{xcolor}         
\usepackage{algorithm}

\usepackage{graphicx}
\usepackage{amsmath}
\usepackage{caption}
\usepackage{graphicx}
\usepackage{multirow}
\usepackage{subcaption}
\usepackage[table]{xcolor}
\usepackage{hyperref}

\usepackage[preprint]{icml2026}

\usepackage{amsmath}
\usepackage{amssymb}
\usepackage{mathtools}
\usepackage{amsthm}

\usepackage[capitalize,noabbrev]{cleveref}

\theoremstyle{plain}

\theoremstyle{definition}

\theoremstyle{remark}

\usepackage[textsize=tiny]{todonotes}

\icmltitlerunning{Factual and Edit-Sensitive Graph-to-Sequence Generation via Graph-Aware Adaptive Noising}

\begin{document}

\twocolumn[
  \icmltitle{Factual and Edit-Sensitive Graph-to-Sequence Generation\\ via Graph-Aware Adaptive Noising}



  \icmlsetsymbol{equal}{*}

 \begin{icmlauthorlist}
    \icmlauthor{Aditya Hemant Shahane}{iitd}
    \icmlauthor{Anuj Kumar Sirohi}{iitd}
    \icmlauthor{Tanmoy Chakraborty}{iitd}\\
    \icmlauthor{Prathosh A P}{iisc,latent}
    \icmlauthor{Sandeep Kumar}{iitd}
\end{icmlauthorlist}

\icmlaffiliation{iitd}{Department of Electrical Engineering, Indian Institute of Technology Delhi, New Delhi, India}
\icmlaffiliation{iisc}{Indian Institute of Science, Bengaluru, India}
\icmlaffiliation{latent}{Latentforce.ai}

\icmlcorrespondingauthor{Aditya Hemant Shahane}{eez248435@iitd.ac.in}
\vspace{15pt}
]



\printAffiliationsAndNotice{}  

\begin{abstract}
Fine-tuned autoregressive models for graph-to-sequence generation (G2S) often struggle with factual grounding and edit sensitivity. To tackle these issues, we propose a non-autoregressive diffusion framework that generates text by iterative refinement conditioned on an input graph, named as \textbf{D}iffusion \textbf{L}anguage \textbf{M}odel \textbf{for} \textbf{G}raphs~(\texttt{DLM4G}). By aligning graph components (entities/relations) with their corresponding sequence tokens, \texttt{DLM4G} employs an adaptive noising strategy. The proposed strategy uses per-token denoising error as a signal to adaptively modulate noise on entity and relation tokens, improving preservation of graph structure and enabling localized updates under graph edits. Evaluated on three datasets, \texttt{DLM4G} consistently outperforms competitive G2S diffusion baselines trained on identical splits across both surface-form and embedding-based metrics. \texttt{DLM4G} further exceeds fine-tuned autoregressive baselines up to 12$\times$ larger (e.g., T5-Large) and is competitive with zero-shot LLM transfer baselines up to 127$\times$ larger. Relative to the strongest fine-tuned PLM baseline, \texttt{DLM4G} improves factual grounding (FGT@0.5) by \textbf{+5.16\%} and edit sensitivity (ESR) by \textbf{+7.9\%}; compared to the best diffusion baseline, it yields gains of \textbf{+3.75\%} in FGT@0.5 and \textbf{+23.6\%} in ESR. We additionally demonstrate applicability beyond textual graphs through experiments on molecule captioning, indicating the method’s generality for scientific G2S generation.
\vspace{-8mm}
\end{abstract}

\section{Introduction}

Graphs are widely used to represent relational data across domains such as social platforms, biological systems, and recommender systems \cite{wang2021graphlearningbasedrecommender, 10.5555/3692070.3694214}. Natural-language verbalization of structured graph inputs is increasingly important, both for communicating graph content and as textual context for LLM-based reasoning \cite{skianis2024graphreasoninglargelanguage}. This motivates \textit{graph-to-sequence} (G2S) generation, where the goal is to produce a coherent output sequence conditioned on an input graph \cite{fatemi2024talk}. Real-world instances include (i) \emph{scientific captioning}, where molecular or protein graphs are verbalized into concise natural-language descriptions \cite{kim2025grapht5unifiedmoleculargraphlanguage}, and (ii) \emph{knowledge-graph based QA}, where relevant KG subgraphs are verbalized to support multi-hop reasoning \cite{wu2023retrieverewriteanswerkgtotextenhancedllms}.

Early G2S approaches paired graph encoders with sequence decoders to inject structural bias explicitly \cite{ribeiro-etal-2020-modeling, schmitt2021}. More recent work shows that autoregressive pre-trained language models (PLMs), when fine-tuned for G2S, can achieve strong performance on surface-overlap metrics such as BLEU and chrF++ even without graph-specific inductive biases \cite{ribeiro-etal-2021-investigating}. However these metrics do not enforce coverage of the input graph: scores can remain high even when generated text contains \emph{factual omissions} (missing entities/relations) or \emph{hallucinations} (introducing unsupported facts). In addition, G2S systems are often deployed in interactive or evolving settings where the input graph is edited (e.g., adding/removing a neighbor, changing an entity, modifying a relation). In such cases, a desirable property is \emph{edit sensitivity}: small, localized graph edits should induce predictable and localized changes in the generated text.

One contributing factor for these failures is the left-to-right commitment of autoregressive decoding: once early tokens are generated, subsequent tokens are constrained by the prefix, making it difficult to revise earlier decisions \cite{li2022diffusionlmimprovescontrollabletext, gong2023diffuseqsequencesequencetext}. In contrast, non-autoregressive diffusion language models generate by \emph{iterative refinement}, repeatedly denoising a corrupted sequence, which naturally supports global self-correction \cite{gong2025scaling, venkatraman2025amortizingintractableinferencediffusion}. While iterative refinement typically incurs higher inference cost than autoregressive decoding, it provides a built-in mechanism for revising earlier decisions and improving global consistency with the input graph. Yet standard diffusion schedules are typically \emph{data-agnostic}, corrupting all tokens uniformly \cite{ho2020denoisingdiffusionprobabilisticmodels}. For G2S, uniform corruption can be misaligned with the input structure: tokens aligned to entities and relations may require different corruption/denoising than purely functional or syntactic tokens, and treating them identically can degrade factual grounding.

To address these gaps, we propose \textbf{D}iffusion \textbf{L}anguage \textbf{M}odel \textbf{for} \textbf{G}raphs~(\texttt{DLM4G}), a framework that leverages diffusion-based iterative refinement while preserving graph facts during corruption. Motivated by prior G2S findings that PLMs provide strong semantics but lack explicit structural guarantees \cite{ribeiro-etal-2021-investigating}, we retain a PLM backbone and re-introduce graph structure through a soft inductive bias in the noising process, rather than imposing hard architectural constraints (e.g., graph encoders with permutation equivariance).
The key idea is to align graph components (entities and relations) with their corresponding sequence tokens, and implement a \emph{graph-aware adaptive noising schedule} that uses each aligned token's denoising error as a difficulty signal: tokens that are harder to recover (typically entity/relation tokens) are corrupted less, while easier tokens are noised uniformly. This biased corruption preserves graph information during refinement and improves factual grounding, while also promoting localized updates when the input graph is edited.
Finally, we complement modeling improvements with task-grounded evaluation by introducing metrics that directly measure (i) factual grounding and (ii) edit sensitivity, moving beyond surface overlap to quantify whether generated text faithfully reflects the input graph and responds predictably to local edits.

\textbf{Contributions.}
To sum up, our main contributions are:
(1) a novel graph-aware adaptive noising schedule for diffusion-based G2S to improve factual grounding;
(2) strong performance on three diverse benchmarks across surface and embedding-based metrics;
(3) two task-grounded metrics for factual grounding and edit sensitivity; and
(4) an extension to molecule captioning, demonstrating applicability to scientific G2S generation.

\section{Background and Preliminaries}\label{sec:background}


\subsection{Related Work}
\textbf{Graph-to-Sequence Learning.}
G2S has progressed from (i) template-based systems \cite{kasner-dusek-2022-neural, vejvar-fujimoto-2023-aspiro}, to (ii) neural encoder--decoders with learned graph representations \cite{ribeiro-etal-2020-modeling, schmitt2021}, and (iii) fine-tuned transformer-based models that achieve competitive performance on standard surface metrics (BLEU, chrF++, METEOR) even without explicit graph-specific inductive biases \cite{ribeiro-etal-2021-investigating, NIPS2017_3f5ee243}.
This evolution motivates revisiting whether standard-metric performance also implies faithful realization of the input graph.

\textbf{PLMs for Graph Verbalization.}
A common strategy is to serialize the input graph (e.g., a KG) into a sequence of relational triples using special markers such as \texttt{[HEAD]}, \texttt{[REL]}, \texttt{[TAIL]}, and \texttt{[SEP]} (see Section~\ref{Graph_representation}). This enables direct use of standard encoder--decoder Transformers, but it gives up strict permutation invariance, a trade-off we revisit in Limitations~\ref{sec:limitations}. Recent work also highlights challenges in (i) aligning graph elements to text and (ii) representing multi-level semantics across nodes, edges, and subgraphs \cite{zhu2025llmgnngraphvocabulary}. While PLM-based approaches can be fluent, they remain constrained by left-to-right decoding, which can amplify early commitments and make local input edits propagate unpredictably \cite{ gong2023diffuseqsequencesequencetext}.

\textbf{Diffusion Models for Conditional Generation.}
Diffusion models provide a non-autoregressive alternative that generates by iterative refinement. In text, early approaches such as Diffusion-LM \cite{li2022diffusionlmimprovescontrollabletext} and Analog Bits \cite{chen2023analog} explored controllable generation in continuous representations, while DIFFUSEQ \cite{gong2023diffuseqsequencesequencetext} introduced sequence-to-sequence conditioning for diffusion language models. \texttt{DLM4G} builds on this line by conditioning explicitly on a serialized knowledge graph and introducing graph-aware adaptive noising to better preserve and recover graph-aligned components.

\textbf{Molecule Captioning.}
Graph-to-text generation is also central in scientific domains such as molecule and protein captioning \cite{edwards-etal-2022-translation, LIU2024108073, kim2025grapht5unifiedmoleculargraphlanguage, fei2025prot2textv2proteinfunctionprediction}. We include molecule captioning as an additional evaluation setting for graph-conditioned diffusion-based G2S. Table~\ref{tab:comparison} compares prior paradigms with \texttt{DLM4G}.

\subsection{Preliminaries: Denoising Diffusion Models}

DDPMs learn a data distribution over continuous variables, optionally conditioned on context $\mathbf{c}$, denoted $p(\mathbf{z}_0 \mid \mathbf{c})$. They consist of a fixed forward diffusion process and a learned reverse denoising process.

\textbf{Forward process.}
A standard DDPM forward process corrupts clean data $\mathbf{z}_0$ through a Markov chain with noise-schedule coefficients $\{\alpha_t\}_{t=1}^T$ controlling signal decay. This yields a closed-form for sampling a noised state $\mathbf{z}_t$ at any timestep $t$: $\mathbf{z}_t = \sqrt{\bar{\alpha}_t}\,\mathbf{z}_0 + \sqrt{1 - \bar{\alpha}_t}\,\boldsymbol{\epsilon},$
$\text{with} \quad
    \bar{\alpha}_t = \prod_{s=1}^t \alpha_s
    \ \text{and} \
    \boldsymbol{\epsilon}\sim\mathcal{N}(\mathbf 0,\mathbf I)$. Standard diffusion models typically use a fixed, data-agnostic (isotropic) schedule.

\textbf{Reverse process with conditional denoising.}
The reverse process learns to recover the clean data $\mathbf{z}_0$ from pure noise $\mathbf{z}_T \sim \mathcal{N}(\mathbf 0,\mathbf I)$. Let $q(\mathbf{z}_{1:T}\mid \mathbf{z}_0)$ denote the forward diffusion distribution. The learned reverse model defines a Markov chain $p_\theta(\mathbf{z}_{0:T}\mid \mathbf{c})$ where each transition
$p_\theta(\mathbf{z}_{t-1} \mid \mathbf{z}_t, \mathbf{c})$ is Gaussian with parameters predicted by a neural network $\mathcal{M}_\theta(\mathbf{z}_t,t,\mathbf{c})$ (e.g., via a parameterization of $\boldsymbol{\mu}_\theta$ and $\boldsymbol{\Sigma}_\theta$).
The parameters $\theta$ are optimized by maximizing a variational lower bound (VLB) on the conditional log-likelihood:
\vspace{-3mm}
\begin{equation}
\label{eq:vlb}
\resizebox{\columnwidth}{!}{$
\begin{aligned}
\mathcal{L}_{\text{vlb}}
= \mathbb{E}_{q}\Big[
\underbrace{-\log p_\theta(\mathbf{z}_0 \mid \mathbf{z}_1, \mathbf{c})}_{\text{Reconstruction }(L_0)}
 + \sum_{t=2}^{T}
\underbrace{
D_{\mathrm{KL}}\!\Big(
q(\mathbf{z}_{t-1} \mid \mathbf{z}_t, \mathbf{z}_0)\ \|\ 
p_\theta(\mathbf{z}_{t-1} \mid \mathbf{z}_t, \mathbf{c})
\Big)
}_{\text{Denoising matching }(L_{t-1})}
\\
+ \underbrace{
D_{\mathrm{KL}}\!\Big(
q(\mathbf{z}_T \mid \mathbf{z}_0)\ \|\ p(\mathbf{z}_T)
\Big)
}_{\text{Prior matching }(L_T)}
\Big].
\end{aligned}
$}
\end{equation}

While tractable, direct optimization of the full VLB is often unstable~\citep{li2022diffusionlmimprovescontrollabletext}; in later sections we show the modified objective with graph-aware, component-wise noising.
\begin{figure*}[t!]
    \centering
    \includegraphics[width=0.8\linewidth, height = 4cm]{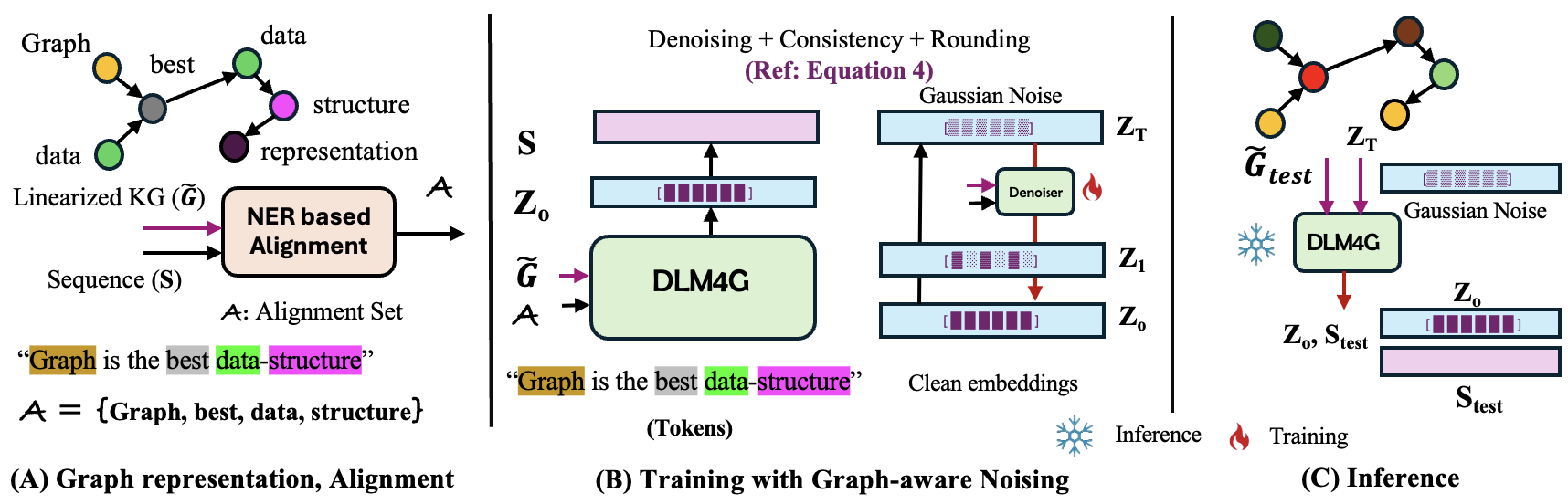}
    \caption{\texttt{DLM4G} framework:
(A) Graph-Sequence alignment set \{$\mathcal{A}$\}, obtains the aligned tokens;
(B) The model is trained with a graph-aware noising schedule
(C) Trained \texttt{DLM4G} samples output sequence conditioned on graph.}
\vspace{-5mm}
    \label{dlm4g}
\end{figure*}

\vspace{-5mm}
\section{The \texttt{DLM4G} Methodology}
\label{sec:method}

\subsection{Problem Setting: Graph-to-Sequence Generation}
\label{sec:problem}

Let $\mathcal{G}=(\mathcal{V},\mathcal{E},\mathbf{X})$ denote an input graph, where
$\mathcal{V}=\{v_1,\dots,v_n\}$ is the node set, $\mathcal{E}$ is the edge set, and
$\mathbf{X}=\{x_i\}_{i=1}^n$ with $x_i\in\mathbb{R}^d$ are node features.
Our goal is \emph{graph-to-sequence} (G2S) generation: learning a conditional distribution
$p_\theta(\mathbf{S}\mid \mathcal{G})$ over output sequences
$\mathbf{S}=(s_1,\dots,s_N)$, where $s_i\in\mathcal{W}$ and $\mathcal{W}$ is the vocabulary.
In practice, variable-length targets are padded/truncated to a maximum length $N$.
We further define $\mathcal{M}_\theta$ as a parameterized model that induces
$p_\theta(\mathbf{S}\mid \mathcal{G})$.
Our objective is maximum likelihood learning under a dataset $\mathcal{D}$: $\theta^\star = \arg\max_\theta \;\mathbb{E}_{(\mathcal{G},\mathbf{S})\sim\mathcal{D}}
\big[\log p_\theta(\mathbf{S}\mid \mathcal{G})\big].$

\textbf{Knowledge graphs as a special case.}
A knowledge graph (KG) is a relational graph $\mathcal{G}_{\mathrm{KG}}=(\mathcal{V},\mathcal{R},\mathcal{T})$
defined by a set of entities $\mathcal{V}$, relation types $\mathcal{R}$, and directed triples
$\mathcal{T}\subseteq \mathcal{V}\times\mathcal{R}\times\mathcal{V}$.
We denote triples as $(h,r,t)\in\mathcal{T}$.
In this work, we instantiate our G2S framework on KGs via a standard serialization into a token sequence
$\mathbf{G}=\textsc{Serialize}(\mathcal{T})$ (Section~\ref{Graph_representation}),
but the formulation above remains generalized.
\vspace{-6mm}
\subsection{\texttt{DLM4G}: Overview}
\label{sec:dlm4g_overview}

\texttt{DLM4G} is a \emph{graph-conditioned diffusion language model} that generates a sequence by iterative refinement.
We adapt conditional DDPMs (Section~\ref{sec:background}) to G2S by operating on a continuous sequence representation.
Let the clean latent sequence be
$\mathbf{z}_0 = g_\Phi(\mathbf{S})\in\mathbb{R}^{N\times d}$,
where $g_\Phi$ is a learnable embedding/projection layer.
We condition generation on a tokenized graph representation $\mathbf{G}$, obtained by serializing the input graph as a sequence of directed triples (full details in
Section~\ref{Graph_representation}): $\mathbf{G}
=
\big\langle
\texttt{[HEAD]}\ h\ \texttt{[REL]}\ r\ \texttt{[TAIL]}\ t\ \texttt{[SEP]}
\big\rangle_{(h,r,t)\in\mathcal{T}}$
where $\mathcal{T}$ denotes the directed edge set expressed as triples, and \texttt{[SEP]} is the delimiter.
Finally, our conditional generation model is denoted by
$\mathcal{M}_\theta(\mathbf{z}_t,t,\mathbf{G})$.

The key methodological difference in \texttt{DLM4G} is the \emph{graph-aware adaptive noising schedule} that allocates different noise levels to
graph-aligned tokens (entity/relation mentions) versus unaligned tokens ( functional/syntactic content).
This induces a soft structural bias in the diffusion trajectory while retaining the semantic capabilities of a PLM backbone. Additional details are in Section~\ref{sec:graph_aware_noising}
\vspace{-3mm}
\subsection{Graph-Aware Adaptive Noising}
\label{sec:graph_aware_noising}

\textbf{Motivation.}
Standard diffusion models use data-agnostic schedules that corrupt all token positions uniformly
\citep{ nichol2021improveddenoisingdiffusionprobabilistic}.
For G2S generation, uniform corruption ignores that entity/relation tokens are more critical for faithfulness than syntactic/functional tokens. This makes entity/relation mentions harder to recover at high-noise
timesteps, increasing omissions or hallucinations.

\textbf{Rationale.}
We introduce a \emph{graph-aware adaptive noising schedule} that assigns different noise levels
to graph-aligned vs.\ unaligned token positions. The key signal is a token-wise denoising error
$\ell_t^{\,i}$ for an aligned token $i$ at timestep $t$. Larger $\ell_t^{\,i}$ indicates that
the model struggles to recover that token under the current noise level, suggesting it should retain
a higher signal-to-noise ratio (higher $\bar{\alpha}$) throughout diffusion. We periodically estimate these
errors during training (every $K_{\text {up}}$ steps) and update token-wise cumulative schedules accordingly.

\textbf{Padding and indexing.}
All targets are padded/truncated to a fixed length $N$. Throughout, $i\in\{1,\dots,N\}$ denotes a token position within a single training example;
diffusion is applied only to non-\texttt{[PAD]} positions, and padding tokens are excluded from the forward
process and losses.

\textbf{Graph--sequence alignment (Training).}\label{alignment}
To identify which output token positions correspond to graph elements, we construct a training-only alignment set
$\mathcal{A}\subseteq\{1,\dots,N\}$ containing only non-\texttt{[PAD]} positions whose tokens mention KG entities/relations.
We use a one-time offline pipeline that (i) enumerates entity aliases, (ii) detects mentions, and (iii) links mentions
to KG entities to resolve ambiguity. This alignment is used only during training; details are deferred to
Appendix~\ref{app:alignment-analysis}.

\noindent\textbf{Graph-aware vs.\ baseline noising.}
We apply graph-aware noising only to aligned positions $i\in\mathcal{A}$, while keeping unaligned positions
$i\notin\mathcal{A}$ on the baseline cumulative schedule $(\bar{\alpha}_t)_{t=1}^T$.
We instantiate the forward process in Eq.~(1)  by replacing
$\bar{\alpha}_t$ with $\bar{\alpha}_t^{\,i}$; \texttt{[PAD]} positions are not considered during noising.
The procedure has two stages: Stage~1 estimates token-wise denoising difficulty for $i\in\mathcal{A}$, and Stage~2
constructs token-wise schedules $\{\bar{\alpha}^{\,i}_{t,\mathrm{new}}\}_{t=1}^T$ (Algorithm~\ref{alg:graph_aware_training}).

\textbf{Stage 1: Estimating token-wise denoising difficulty.}\\
For each aligned position $i\in\mathcal{A}$ and timestep $t\in\{1,\dots,T\}$, we define
\begin{equation}
\label{eq:token_difficulty}
\ell_t^{\,i}
=
\mathbb{E}_{\mathbf{z}_t \sim q(\mathbf{z}_t \mid \mathbf{z}_0;\,\{\bar{\alpha}^{\,j}_t\}_{j=1}^N)}
\Big\|
\mathcal{M}_\theta(\mathbf{z}_t,t,\mathbf{G})^{(i)}
-
\mathbf{z}_0^{(i)}
\Big\|^2,
\end{equation}
where $\mathcal{M}_\theta(\cdot)^{(i)}$ denotes the model output at position $i$ and
$q(\mathbf{z}_t\mid \mathbf{z}_0;\{\bar{\alpha}^{\,j}_t\}_{j=1}^N)$ uses the current token-wise schedules
(initially the baseline; after updates, the learned schedules on $i\in\mathcal{A}$).
Aggregating over aligned token occurrences in the training set results in a difficulty profile
$(\ell_1^{\,i},\dots,\ell_T^{\,i})$.

Since $(\ell_t^{\,i})_{t=1}^T$ need not be monotone in $t$, it cannot be used directly to construct a valid diffusion schedule.
In Stage~2 the difficulty profile is mapped to \emph{per-step} noising coefficients to reconstruct a valid
token-wise cumulative schedule
$1=\bar{\alpha}^{\,i}_0 \ge \bar{\alpha}^{\,i}_1 \ge \cdots \ge \bar{\alpha}^{\,i}_T>0$. The noise schedules are updated every $K_{\text {up}}$ steps and loss-to-noise mapping is performed
over non-overlapping windows of length $K_{\text {win}}$.

\textbf{Stage 2: Token-wise cumulative schedule.}\\
Let $(\bar{\alpha}_t)_{t=1}^T$ denote a baseline cumulative schedule with $\bar{\alpha}_t\in(0,1)$ and $\bar{\alpha}_0=1$.
Define baseline per-step coefficients $\alpha_t^{\mathrm{base}}=\frac{\bar{\alpha}_t}{\bar{\alpha}_{t-1}}$ and
$\beta_t^{\mathrm{base}}=1-\alpha_t^{\mathrm{base}}$ for $t=1,\dots,T$.
For each aligned token $i\in\mathcal{A}$, we construct token-wise per-step coefficients
$\{\alpha_{t,i}\}_{t=1}^T$ such that larger denoising difficulty implies less noising
(larger $\alpha_{t,i}$), and then reconstruct a valid cumulative schedule
$\bar{\alpha}^{\,i}_{t,\mathrm{new}}=\prod_{k=1}^t \alpha_{k,i}$.

To enforce \emph{higher error $\Rightarrow$ higher signal}, we perform the loss-to-noise mapping at the diffusion-window level.
Let $\tau>0$ be a constant and let
$W_m=\{(m-1)K_{\mathrm{win}}+1,\dots,\min(mK_{\mathrm{win}},T)\}$ denote the $m$-th diffusion window.
Let $M=\left\lceil \frac{T}{K_{\mathrm{win}}}\right\rceil$ be the number of windows, and define
$t_m=\min(mK_{\mathrm{win}},T)$ and $t_{m-1}=(m-1)K_{\mathrm{win}}$.
For each token $i$, define the window-averaged difficulty $\tilde{\ell}_m^{\,i}=\frac{1}{|W_m|}\sum_{t\in W_m}\ell_t^{\,i}$.
Define the window extrema over aligned tokens
$\ell_m^{\min}=\min_{j\in\mathcal{A}}\tilde{\ell}_m^{\,j}$ and
$\ell_m^{\max}=\max_{j\in\mathcal{A}}\tilde{\ell}_m^{\,j}$.
For each window $W_m$, define a loss-to-coefficient mapping function:
\begin{equation}
\label{eq:window_mapping}
\Psi_m(x)
=
\alpha_{t_m}^{\mathrm{base}}
+
\frac{x-\ell_m^{\min}}
{\ell_m^{\max}-\ell_m^{\min}+\tau}
\big(\alpha_{t_{m-1}}^{\mathrm{base}}-\alpha_{t_m}^{\mathrm{base}}\big).
\end{equation}
We then set the window-wise per-step coefficient
$\tilde{\alpha}_{m,i}=\mathrm{clip}\!\big(\Psi_m(\tilde{\ell}_m^{\,i}),\,\alpha_{\min},\,1\big)$,
and set $\alpha_{t,i}=\tilde{\alpha}_{m,i}$ for all $t\in W_m$.
For the first window $W_1$, we initialize $\alpha_{t,i}=\alpha_t^{\mathrm{base}}$.
Finally, define $\beta_{t,i}=1-\alpha_{t,i}$ and reconstruct
$\bar{\alpha}^{\,i}_{t,\mathrm{new}}=\prod_{k=1}^{t}\alpha_{k,i}$,
with $\bar{\alpha}^{\,i}_{0,\mathrm{new}}=1$.
Unaligned tokens retain the baseline schedule:
$\bar{\alpha}^{\,i}_{t,\mathrm{new}}=\bar{\alpha}_t$ for $i\notin\mathcal{A}$.

\subsection{Training Objective}
\label{sec:training_objective}
We derive our training objective from the conditional VLB (Eq.~\ref{eq:vlb}).
Rather than optimizing the full VLB directly, we adopt an $\mathbf{z}_0$-prediction parameterization and train
$\mathcal{M}_\theta$ to predict the clean latent $\mathbf{z}_0$ from $\mathbf{z}_t$ at each timestep.
To map the continuous latent back to discrete tokens, we use a learned rounding distribution
$\tilde{p}_\Phi(\mathbf{S}\mid \mathbf{z}_0)$.
The resulting end-to-end objective is:
\begin{equation}
\label{eq:e2e_simple}
\resizebox{0.8\columnwidth}{!}{$
\begin{aligned}
\mathcal{L}_{\text{e2e-simple}}(\mathbf{S})
&= \mathbb{E}_{q(\mathbf{z}_{1:T}\mid \mathbf{z}_0)}\Bigg[
\sum_{t=2}^{T}
\underbrace{\big\|\mathcal{M}_\theta(\mathbf{z}_t, t, \mathbf{G}) - \mathbf{z}_0 \big\|^2}_{\text{Denoising}}
\\
&\quad + \underbrace{\big\|g_\Phi(\mathbf{S}) - \mathcal{M}_\theta(\mathbf{z}_1, 1, \mathbf{G})\big\|^2}_{\text{Consistency}}
- \underbrace{\log \tilde{p}_\Phi(\mathbf{S} \mid \mathbf{z}_0)}_{\text{Rounding}}
\Bigg].
\end{aligned}
$}
\end{equation}
Here $g_\Phi(\mathbf{S})\in\mathbb{R}^{N\times d}$ denotes the token embedding lookup, i.e.,
$\mathbf{z}_0=g_\Phi(\mathbf{S})$ with $(\mathbf{z}_0)_i=\mathbf{e}_{s_i}$.
The rounding distribution factorizes across positions,
$\tilde{p}_\Phi(\mathbf{S}\mid \mathbf{z}_0)=\prod_{i=1}^{N}\tilde{p}_\Phi(s_i\mid \mathbf{z}_{0,i})$,
where
$\tilde{p}_\Phi(s_i\mid \mathbf{z}_{0,i})=\mathrm{Softmax}(\mathbf{W}\mathbf{z}_{0,i})_{s_i}$
for a learned projection $\mathbf{W}\in\mathbb{R}^{|\mathcal{V}|\times d}$
(optionally tied to the embedding matrix).
During training we evaluate this term with teacher forcing (using the gold $s_i$); at inference we decode from
$\tilde{p}_\Phi(\cdot\mid \hat{\mathbf{z}}_0)$.
The forward noising samples $\mathbf{z}_t\sim q(\mathbf{z}_t\mid \mathbf{z}_0)$ are generated using the
token-wise schedule from Section~\ref{sec:graph_aware_noising}.
We provide the full VLB derivation and implementation details in Appendix~\ref{sec:derivation_appendix}.
\begin{figure}[h!]
    \centering
\includegraphics[width=0.65\linewidth, height = 3cm]{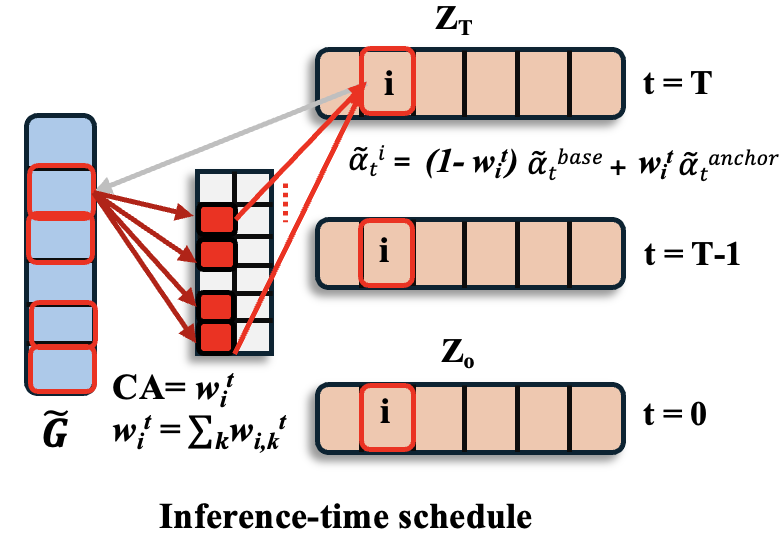}
    \caption{\textbf{Inference-time schedule.} We aggregate decoder-to-graph cross-attention mass from token position $i$ to the serialized graph tokens $\mathbf{G}$ into a weight $w_i^t$, and blend baseline and anchor cumulative schedules to obtain $\bar{\alpha}_t^{\,i}$.}
    \label{fig:inference_schedule}
\end{figure}
\vspace{-5mm}
\subsection{Inference-Time Schedule}
\label{sec:inference_schedule}

At test time, the alignment set $\mathcal{A}$ is unavailable. We therefore use decoder-to-graph cross-attention
as a proxy for whether a generated token position is graph-aligned (Figure~\ref{fig:inference_schedule}). Let $\bar{\alpha}_t^{\mathrm{base}}$ denote the
baseline schedule, and let $\bar{\alpha}_t^{\mathrm{anchor}}$ be an anchor schedule defined as the average of the
learned schedules $\bar{\alpha}^{\,i}_{t,\mathrm{new}}$ over aligned (non-\texttt{[PAD]}) token occurrences in the training
set (and fixed at test time). At denoising step $t$, for each (non-\texttt{[PAD]}) position $i$ we compute $w_i^t\in[0,1]$
as the total normalized cross-attention mass from $i$ to the serialized graph tokens $\mathbf{G}$ (averaged across heads). We then set $\bar{\alpha}_t^{\,i}
=
(1-w_i^t)\,\bar{\alpha}_t^{\mathrm{base}}
+
w_i^t\,\bar{\alpha}_t^{\mathrm{anchor}}.$
Positions with larger $w_i^t$ are noised more conservatively, providing an inference-time approximation to the
training-time graph-aware schedule.
\begin{table*}[t!]
\caption{Performance of \texttt{DLM4G} compared with (i) finetuning and (ii) zero-shot evaluation paradigms.}
  \centering
  \small
  \resizebox{0.7\textwidth}{!}{
    \begin{tabular}{lcccccccccc}
      \toprule
      \textbf{Model} & \textbf{\#P}
        & \multicolumn{3}{c}{\textbf{WikiOFGraph}}
        & \multicolumn{3}{c}{\textbf{GenWiki}}
        & \multicolumn{3}{c}{\textbf{TekGEN}} \\
      \cmidrule(lr){3-5} \cmidrule(lr){6-8} \cmidrule(lr){9-11}
      & & \textbf{B} & \textbf{CrF++} & \textbf{M}
        & \textbf{B} & \textbf{CrF++} & \textbf{M}
        & \textbf{B} & \textbf{CrF++} & \textbf{M} \\
      \midrule

      \multicolumn{11}{l}{\textcolor{blue}{\textit{\# Pretrain}}} \\
      \rowcolor{white!70!yellow}
      \texttt{DLM4G}‑1.o       & 50M  & \underline{0.619} & \underline{0.823} & \underline{0.688}
                              & \underline{0.401} & \underline{0.663} & \underline{0.527}
                              & \underline{0.247} & \underline{0.493} & \underline{0.375} \\
      \rowcolor{white!70!yellow}
      \texttt{DLM4G}‑2.o       & 63M  & \textbf{0.654} & \textbf{0.844} & \textbf{0.791}
                              & \textbf{0.469} & \textbf{0.748} & \textbf{0.574}
                              & \textbf{0.253} & \textbf{0.522} & \textbf{0.414} \\
      \rowcolor{white!70!gray}
      \%Gain          & \textbf{x1.3\(\downarrow\)}  & \textbf{+5.7\%} & \textbf{+2.5\%} & \textbf{+14.9\%}
                              & \textbf{+16.9\%} & \textbf{+12.8\%} & \textbf{+8.9\%}
                              & \textbf{+2.4\%} & \textbf{+5.9\%} & \textbf{+10.4\%} \\

      \multicolumn{11}{l}{\textcolor{blue}{\textit{\# Finetune}}} \\
      GPT‑2 (S)      & 124M & 0.166 & 0.428 & 0.487 & 0.280 & 0.465 & 0.435 & 0.226 & 0.358 & 0.208 \\
      GPT‑2 (B)      & 355M & 0.285 & 0.572 & 0.490 & 0.312 & 0.470 & 0.425 & \underline{0.228} & 0.366 & 0.211 \\
      T5   (S)       &  60M & 0.385 & 0.688 & 0.471 & 0.227 & 0.495 & \underline{0.447} & 0.189 & 0.352 & 0.203 \\
      T5   (L)       & 770M & \textbf{0.658} & \underline{0.807} & \underline{0.516}
                               & \underline{0.361} & \underline{0.567} & 0.338
                               & 0.199 & \underline{0.370} & \underline{0.211} \\
      \rowcolor{white!70!yellow}
      \texttt{DLM4G}-2.o & 63M  & \underline{0.654} & \textbf{0.844} & \textbf{0.791}
                               & \textbf{0.469} & \textbf{0.748} & \textbf{0.574}
                               & \textbf{0.253} & \textbf{0.522} & \textbf{0.414} \\
      \rowcolor{white!70!gray}
      \%Gain      & \textbf{x12\(\uparrow\)}
                               & 0.0\% & \textbf{+4.5\%} & \textbf{+53.3\%}
                               & \textbf{+29.9}\% & \textbf{+31.9\%} & \textbf{+28.4\%}
                               & \textbf{+10.9\%} & \textbf{+41.1\%} & \textbf{+96.2\%} \\

      \addlinespace
      \multicolumn{11}{l}{\textcolor{blue}{\textit{\# Zero‑shot}}} \\
      LLaMa-3  & 8B   & 0.622     & 0.801     & {0.781}     & {0.461}    & 0.709     & 0.510     & 0.176     & 0.341     & 0.251     \\
            
      Qwen2.5   & 7B    & 0.622     & 0.681     & 0.743    & {0.461}     & 0.697     & 0.501     & \underline{0.182}     & 0.312     & 0.234     \\

      DeepSeek   & 7B    & 0.633     & 0.809    & 0.752     & 0.391     & 0.688     & \underline{0.533}     & 0.121     & \underline{0.345}     & 0.256     \\
      GPT‑o4-mini   & 8B    & \underline{0.648}     & \textbf{0.847}     & \underline{0.783}     & \underline{0.464}     & {0.734}     & 0.471     & 0.121    & 0.327     & \underline{0.277}     \\
      \rowcolor{white!70!yellow}
      \texttt{DLM4G}-2.o & 63M  & \textbf{0.654} & \underline{0.844} & \textbf{0.791}
                               & \textbf{0.469} & \textbf{0.748} & \textbf{0.574}
                               & \textbf{0.253} & \textbf{0.522} & \textbf{0.414}\\
      \rowcolor{white!70!gray}
      \%Gain      & \textbf{x127\(\uparrow\)}  & 0.0\%     & 0.0\%     & \textbf{+1.0\%}     & \textbf{+1.1\%}     & \textbf{+2.1\%}     &\textbf{ +7.7\%}     & \textbf{+39.0\%}     & \textbf{+51.3\%}     & \textbf{+49.5\%}    \\
      
      \bottomrule
    \end{tabular}}
  \vspace{-3mm}
  \label{main results}
\end{table*}

\vspace{-3mm}
\section{Experiments}
\subsection{Experimental Setup}
\textbf{Model Architecture}:
\texttt{DLM4G} is an  encoder--decoder Transformer that conditions on the serialized KG input (see \textit{graph representation}\S\ref{Graph_representation}). We evaluate two variants: a 6-encoder/6-decoder configuration (\(\approx 50\)M parameters; \texttt{DLM4G}-1.o) and a 6-encoder/9-decoder configuration (\(\approx 63\)M; \texttt{DLM4G}-2.o), both using GeLU activations. Inputs are tokenized with the \texttt{bert-base-uncased} vocabulary~\cite{devlin-etal-2019-bert}; the control tokens \texttt{[HEAD]}, \texttt{[REL]}, \texttt{[TAIL]}, and \texttt{[SEP]} are introduced as learned special tokens with  embeddings. Other components follow the standard Transformer encoder–decoder design.

\textbf{Graph Representation}: \label{Graph_representation}
We represent the set of relational triples $\mathbf{G}$, as a single linearized sequence. This is achieved by serializing each triplet $(h_i, r_{ij}, t_j) \in \mathbf{G}$ into a string format using special tokens, e.g:
\texttt{$\langle$[HEAD] $\texttt{h}_i$ [REL] $\texttt{r}_{ij}$ [TAIL] $\texttt{t}_j$$\rangle$}, and concatenating them with a  separator token \texttt{[SEP]}. For all KG benchmarks (WikiOFGraph, GenWiki, TekGEN), we simply preserve the triple
order provided in the released datasets and do not reorder or subsample triples; this
dataset-defined order serves as our consistent traversal for linearization. We adopt linearization for the following reasons: (i) it plugs into off-the-shelf backbones and decoding stacks, making ablations across baselines directly comparable; (ii) Transformer self-attention can model long-range interactions across the flattened triples, which is important for faithful realization; and (iii) prior work shows strong performance for linearized KG$\rightarrow$text with PLMs, even without graph-specific inductive bias~\citep{ribeiro-etal-2021-investigating,wang2024mgsamultigranularitygraphstructure}. Example \textit{(graph$\rightarrow$sequence)}: \\ Serialized KG ($\mathbf{G}$): \texttt{$\langle$[HEAD] USA [REL] hosted [TAIL] 1994\_FIFA\_World\_Cup$\rangle$ [SEP] $\langle$[HEAD] USA [REL] capital [TAIL] Washington\_D.C.$\rangle$ [SEP] $\langle$[HEAD] 1994\_FIFA\_World\_Cup [REL] top\_scorer [TAIL] Hristo\_Stoichkov$\rangle$}. \\ Corresponding sequence ($\mathbf{S}$): \textit{``The United States hosted the 1994 FIFA World Cup; its capital is Washington, D.C., and the tournament's top scorer was Hristo Stoichkov''}.

\textbf{Datasets}:
We use three datasets for our experiments: (1) \textit{WikiOFGraph} \citep{kim2024ontology}, a dataset for graph-text task; (2) \textit{GenWiki} \citep{jin-etal-2020-genwiki}, an unsupervised dataset of DBpedia graph pairs; and (3) \textit{TekGEN} \citep{agarwal-etal-2021-knowledge}, a dataset generated by verbalizing Wikidata triples. More details are available in Appendix~\ref{App:dataset_summary}.

\textbf{Baselines}: We benchmark \texttt{DLM4G} against four categories: 
(i) \emph{Fine-tuned PLM baselines}, GPT-2 (Small/Base) \citep{mager-etal-2020-gpt} and T5 (Small/Large) \citep{ribeiro-etal-2021-investigating} (trained with the standard 100k-pair G2S finetuning protocol);
(ii) \emph{Zero-shot LLM transfer}, GPT-o4-mini (8B), LLaMa-3-8B (8B), Qwen 2.5 (7B) and DeepSeek (7B) without task-specific finetuning;
(iii) \emph{SOTA task-specific G2S systems}, ReGen on TekGen \citep{dognin-etal-2021-regen} and Ontology-Free \citep{kim2024ontology}, Rule-Based \citep{schmitt-etal-2020-unsupervised}, and Direct-Transfer/Noisy-Supervised \citep{koncel-kedziorski-etal-2019-text} on WikiOFGraph/GenWiki (excluding CycleGT$_{\text{Base}}$ due to non-standard splits \citep{jin-etal-2020-genwiki, guo-etal-2020-cyclegt});
(iv) \emph{Diffusion baselines}, DiffuSeq \citep{gong2023diffuseqsequencesequencetext}, FlowSeq \citep{hu-etal-2024-flow} and SeqDiffuSeq \citep{yuan-etal-2024-text}, adapted to G2S and trained on the same splits as \texttt{DLM4G}.

\textbf{Implementation Details and Evaluation Metrics}: 
We train \texttt{DLM4G} with diffusion process of \(T=2000\) timesteps, using our graph-aware noising schedule, and inputs are tokenized using the \texttt{bert-base-uncased} vocabulary \citep{devlin-etal-2019-bert}. Training uses a peak learning rate of \(10^{-4}\), 10,000 warm-up steps, and a linear decay schedule, with the adaptive noising schedule updated every 20,000 steps. Full implementation details are provided in Appendix~\ref{App:Train},~\ref{App:FSZS}. 
For evaluation, we report BLEU (\textbf{B}) \citep{papineni-etal-2002-bleu}; chrF++ (\textbf{CrF++}) \citep{popovic-2015-chrf}; and METEOR (\textbf{M}) \citep{banerjee2005meteor}. In addition, we include MAUVE (\textbf{MVE}) \citep{pillutla2023mauvescoresgenerativemodels} for distributional similarity and BERTScore-F1 (\textbf{B-F1}) \citep{zhang2020bertscoreevaluatingtextgeneration} as an embedding-based semantic similarity metric.\\
Beyond these, we introduce two task-grounded metrics: Factual Grounding Metric (\textbf{FGT}), and Edit Sensitivity Rate (\textbf{ESR}). Additional details are provided in Section~\ref{facts}.

\subsection{Experimental Results}\label{results}

We evaluate our design using four different methods: (1) full fine-tuning, (2) zero-shot prompting, (3) state-of-the-art (SOTA) benchmarking and (4) Diffusion baselines. Throughout these tests, we carefully balance model size (\textbf{\#P}arameters) with the amount of data (G2S pairs).  For full fine-tuning, we train large models on a dataset of 100,000 graph-to-sequence pairs and test them on a separate set of 1,000 graphs. In the zero-shot evaluation, we use state-of-the-art LLMs without providing any specific training examples. The results across different performance metrics are shown in Table~\ref{main results}. We compare these outcomes against our own pre-trained DLM4G family of small models (approx. 50-63M parameters). These models, trained on an 80/10/10 split, are evaluated on the same test set. A separate SOTA benchmarking table (see Section~\ref{sota}) compares DLM4G's performance against other task-specific models.

\textbf{Model Development and Scaling}: We started by pre-training the \texttt{DLM4G} family. The \texttt{DLM4G}-2.o (63 M \#P) model was the best performer across all three datasets. Increasing the model size by a modest 1.3x (from 50M to 63M parameters) resulted in a significant performance boost of 2.4\% to 16.9\%. This suggests that further scaling \texttt{DLM4G} is a promising direction. 

\textbf{Performance Against Large-Scale Models}: Using our best model, \texttt{DLM4G}-2.o, we benchmarked it against 12-127 times larger baselines. In full fine-tuning tests against baselines like the 770M parameter T5-Large, our model performed better on nearly every metric, with gains up to 96.2\%. Furthermore, in zero-shot comparisons against models approximately 127x larger (including LLaMa-3 and GPT-o4-mini), \texttt{DLM4G}-2.o remained highly competitive and notably outperformed all of them on the TeKGen dataset~(Table~\ref{main results}). 

\textbf{Semantic Evaluation}: To move beyond traditional surface-level metrics and gain a deeper semantic understanding, we also performed experiments using embedding-based metrics. For this analysis, we compare our model against the best-performing autoregressive baselines using the MAUVE score and BERTScore F1. The results are in Table~\ref{tab:embedding-metrics}.

\begin{table}[t!]
    
    \centering
    \caption{\texttt{DLM4G} across embedding based metrics.}
    \label{tab:embedding-metrics}
    \small
    \resizebox{0.9\linewidth}{!}{%
    \begin{tabular}{llcc>{\columncolor{white!70!yellow}}c>{\columncolor{white!70!gray}}c}
        \toprule
        \textbf{Dataset} & \textbf{Metric} 
            & \textbf{T5 (L)} & \textbf{GPT-o4-mini} & \texttt{DLM4G}-2.o & \textbf{\%Gain} \\
        &   & {\textcolor{blue}{\textit{\# Finetune}}} & {\textcolor{blue}{\textit{\# Zero-shot}}} & {\textcolor{blue}{\textit{\# Pretrain}}} & \\
        \midrule
        \multirow{2}{*}{WikiOFGraph}
            & MVE   & 0.980 & \textbf{0.983} & \underline{0.981} & +0.0\% \\
            & B-F1  & 0.926 & \underline{0.960} & \textbf{0.963} & +0.0\% \\
        \midrule
        \multirow{2}{*}{GenWiki}
            & MVE   & \underline{0.852} & 0.811 & \textbf{0.892} & \textbf{+4.7\%} \\
            & B-F1  & 0.812 & \underline{0.865} & \textbf{0.899} & \textbf{+3.9\%}\\
        \midrule
        \multirow{2}{*}{TekGEN}
            & MVE   & \underline{0.803} & 0.751 & \textbf{0.820} & \textbf{+2.1\%} \\
            & B-F1  & \underline{0.789} & 0.652 & \textbf{0.847} & \textbf{+7.3\%} \\
        \bottomrule
    \end{tabular}%
    }
  \vspace{-6mm}
\end{table}

\noindent\textbf{Analysis of Results:}
Table~\ref{tab:embedding-metrics} shows that \texttt{DLM4G}-2.o achieves SoTA performance on GenWiki and TekGEN.
The most significant improvements are on TekGEN, where our model shows a +7.3\% gain in BERTScore F1 over the next best model.
Similarly, on GenWiki, \texttt{DLM4G}-2.o improves the SoTA by +4.7\% on the MAUVE score.
On WikiOFGraph, our model achieves the highest BERTScore F1.
These results demonstrate that \texttt{DLM4G}-2.o, as a compact pre-trained model, generates semantically rich output that moves beyond simple n-gram matching metrics.

\textbf{Summary (Table~\ref{main results},~\ref{tab:embedding-metrics})}: A key takeaway from these results is that a graph-aware pre-training strategy can enable compact models to match, or even surpass, the performance of much larger task-specific and general-purpose LLMs. Finally, for comprehensive evaluation, we benchmark \texttt{DLM4G} against other state-of-the-art models designed for this task.
\begin{table*}[h!]
\caption{Performance of \texttt{DLM4G} compared with diffusion baselines.}
\centering
\small
\resizebox{0.8\textwidth}{!}{
\begin{tabular}{lccccccccccccc}
\toprule
\textbf{Model} & \textbf{\#P}
  & \multicolumn{4}{c}{\textbf{WikiOFGraph}}
  & \multicolumn{4}{c}{\textbf{GenWiki}}
  & \multicolumn{4}{c}{\textbf{TekGEN}} \\
\cmidrule(lr){3-6} \cmidrule(lr){7-10} \cmidrule(lr){11-14}
& &
\textbf{B} & \textbf{M} & \textbf{B-F1} & \textbf{MVE} &
\textbf{B} & \textbf{M} & \textbf{B-F1} & \textbf{MVE} &
\textbf{B} & \textbf{M} & \textbf{B-F1} & \textbf{MVE} \\
\midrule
\multicolumn{14}{l}{\textcolor{blue}{\textit{\# Diffusion Baselines}}} \\
FlowSeq         & 91M  
  & 0.488 & 0.508 & 0.901 & 0.830  
  & 0.133 & 0.387 & 0.855 & 0.672  
  & 0.091 & 0.223 & 0.673 & 0.409 \\
DiffuSeq        & 91M  
  & \underline{0.628} & 0.619 & \underline{0.923} & 0.942  
  & \underline{0.432} & 0.523 & \underline{0.861} & 0.717  
  & 0.154 & 0.198 & 0.797 & 0.725 \\
SeqDiffuSeq     & 50M  
  & 0.616 & 0.649 & \underline{0.923} & 0.947  
  & \underline{0.432} & 0.503 & 0.857 & 0.759  
  & 0.154 & 0.396 & 0.835 & 0.791 \\
\multicolumn{14}{l}{\textcolor{blue}{\textit{\# Ours}}} \\
\rowcolor{white!70!yellow}
\texttt{DLM4G}-1.o & 50M
  & 0.619 & \underline{0.688} & 0.914 & \underline{0.957}
  & 0.401 & \underline{0.527} & 0.837 & \underline{0.822}
  & \underline{0.247} & \underline{0.375} & \underline{0.823} & \underline{0.811} \\
\rowcolor{white!70!yellow}
\texttt{DLM4G}-2.o & 63M
  & \textbf{0.654} & \textbf{0.791} & \textbf{0.963} & \textbf{0.981}
  & \textbf{0.469} & \textbf{0.574} & \textbf{0.899} & \textbf{0.892}
  & \textbf{0.253} & \textbf{0.414} & \textbf{0.847} & \textbf{0.820} \\
\rowcolor{white!70!gray}
\%Gain & \textbf{x1.5\(\uparrow\)}
  & \textbf{+4.1\%} & \textbf{+14.9\%} & \textbf{+4.3\%} & \textbf{+2.5\%}
  & \textbf{+8.5\%} & \textbf{+8.9\%} & \textbf{+4.4\%} & \textbf{+8.5\%}
  & \textbf{+2.4\%} & \textbf{+10.4\%} & \textbf{+2.9\%} & \textbf{+1.1\%} \\
\bottomrule
\end{tabular}}
\label{tab:diffusion_baselines}
\vspace{-4mm}
\end{table*}

\begin{table}[b!]
\vspace{-4mm}
\centering
\small
\setlength{\tabcolsep}{4pt}
\caption{Ablation of noise schedules and mapping function.}
\label{ablation}
\resizebox{0.95\columnwidth}{!}{%
\begin{tabular}{@{}lcccccccc@{}}
\toprule
\texttt{DLM4G} & \textbf{Tokens} & \textbf{$\Psi_i(x)$} & \textbf{B} & \textbf{M} & \textbf{B-F1} & \textbf{MVE} & \textbf{FGT@0.5} & \textbf{ESR} \\
\midrule
\textit{sqrt} (baseline) & All & linear & 0.60 & 0.73 & 0.94 & 0.94 & 0.80 & 0.62 \\
Graph-aware              & All & linear & 0.63 & 0.77 & \textbf{0.97} & \underline{0.98} & 0.83 & 0.67 \\
\rowcolor{white!70!yellow} Graph-aware & $\mathcal{A}$ & linear & \textbf{0.65} & \textbf{0.79} & \underline{0.96} & \textbf{0.98} & \textbf{0.83} & \textbf{0.68} \\
\rowcolor{white!70!yellow} Graph-aware & $\mathcal{A}$ & poly   & 0.61 & 0.72 & 0.93 & 0.96 & 0.81 & 0.64 \\
\rowcolor{white!70!yellow} Graph-aware & $\mathcal{A}$ & cosine & 0.62 & 0.77 & 0.94 & 0.96 & 0.81 & 0.62 \\
\bottomrule
\end{tabular}%
}
\vspace{-6mm}
\end{table}

\textbf{SoTA Benchmarking}:\label{sota} Results in the Table~\ref{full table} confirm that \texttt{DLM4G}-2.o outperforms specialized baselines. On the TekGEN dataset, our model establishes a new SOTA on all five metrics, with performance gains reaching as high as +96.2\% on METEOR. The results are similarly strong on GenWiki, where \texttt{DLM4G}-2.o sets a new SOTA on four of the five metrics and nearly matching the baseline's performance on the final one. Its robust performance across both surface-level and embedding-based metrics highlights the model's ability to generate text that is both lexically accurate and semantically coherent.

\textbf{Diffusion Baselines}: Finally, we evaluate \texttt{DLM4G} against other diffusion-based text generation models (Table~\ref{tab:diffusion_baselines}). Despite being nearly $1.5\times$ smaller than the strongest baseline (91M vs. 63M), \texttt{DLM4G}-2.o demonstrates superior efficiency, consistently outperforming all baselines across every dataset and metric.

\vspace{-3mm}
\subsection{Factual Grounding and Edit Sensitivity}\label{facts}
While the results on established metrics in Section ~\ref{results} demonstrate our model's fluency, these scores are often insufficient for capturing the critical demands of G2S tasks: factual grounding to the source graph and sensitivity to its edits. To address this evaluation gap, we now introduce two novel, task-grounded metrics. To ensure a fair and direct comparison against the baseline results, we conduct this analysis on the WikiOFGraph dataset.

\textbf{Setup and Notations}: For the input KG ($\mathbf{G}$), we extract distinct entities as $\mathcal{U}_{\mathbf{G}} = \{h_i,t_j \mid (h_i, r_{ij}, t_j) \in \mathbf{G}\}$.
For the corresponding generated sequence $\mathbf{S}$, we represent the extracted entities as $\mathcal{U}_{\mathbf{S}}
=\{\,u\mid u\in \mathbf{S}\,\}$. Additionally, we maintain a hallucination set for the output: entities in $\mathbf{S}$ that are not members of $\mathcal{U}_{\mathbf{G}}$ constitute $\mathcal{H}_{\mathbf{S}}$ (with sequence length $N=|\mathbf{S}|$). For the entity and relation extraction, we use the alignment module discussed previously in section~\ref{alignment}. 

\textbf{Factual Grounding Metric (FGT, $\uparrow$)}: FGT measures how precisely the output realizes graph entities, with an optional penalty for out-of-graph mentions. We define
Factual Grounding Metric (FGT) as:
\begin{equation}
\mathcal{F}_{\mathrm{GT}}(\mathbf{G},\mathbf{S})
=
\underbrace{\frac{2\,|\mathcal{U}_{\mathbf{G}}\cap\mathcal{U}_{\mathbf{S}}|}
{|\mathcal{U}_{\mathbf{G}}|+|\mathcal{U}_{\mathbf{S}}|}}_{\text{F1}}
\left(1-\lambda\frac{|\mathcal{H}_{\mathbf{S}}|}{N}\right).
\end{equation}
We report results for $\lambda \in \{0,\;0.5,\;1\}$ and use $\lambda=0.5$ by default, to balance the penalty term.

\textbf{Edit Sensitivity Rate (ESR, $\uparrow$)}: ESR is a precision focused metric. It evaluates whether the edits in graph are realized in its generated sequence. Consider an original pair ($\mathbf{G}, \mathbf{S}$) and an edited pair ($\tilde{\mathcal{G'}}, \mathbf{S'}$). We build $\mathcal{U}_{\mathbf{G}}, \mathcal{U}_{\tilde{\mathcal{G'}}}, \mathcal{U}_{\mathbf{S}}, \mathcal{U}_{\mathbf{S'}}$ as we do in FGT. The graph and text edits (e.g., additions or deletions) are defined as: $\Delta \mathcal{G} \;=\; \big(\mathcal{U}_{\mathbf{G}'} \setminus \mathcal{U}_{\mathbf{G}}\big)\;\cup\;\big(\mathcal{U}_{\mathbf{G}} \setminus \mathcal{U}_{\mathbf{G}'}\big)$ and 
$\Delta \mathcal{T} \;=\; \big(\mathcal{U}_{\mathbf{S}'} \setminus \mathcal{U}_{\mathbf{S}}\big)\;\cup\;\big(\mathcal{U}_{\mathbf{S}} \setminus \mathcal{U}_{\mathbf{S}'}\big)
$. We define Edit Sensitivity Rate (ESR) as: 
\begin{equation}
    \mathrm{\mathcal{E}_{SR}(\mathbf{G},\mathbf{S})} \;=\; \frac{\lvert \Delta \mathcal{G} \cap \Delta \mathcal{T} \rvert}{\lvert \Delta \mathcal{T} \rvert},
\end{equation}
If the text does not change ($|\Delta \mathcal{T}|=0$), set $\mathrm{ESR}=1$ when the graph also does not change ($|\Delta \mathcal{G}|=0$) and $\mathrm{ESR}=0$ when the graph does change ($|\Delta \mathcal{G}|>0$).

To evaluate FGT and ESR, we create edited graphs by randomly substituting a single entity with a plausible alternative from the vocabulary. We then measure whether the output text accurately reflects this specific modification. We compare \texttt{DLM4G} with comparably-size G2S models finetuned on the same task, and report FGT@\{0, 0.5, 1\} and ESR.

\begin{table*}[h!]
\centering
\caption{Performance of \texttt{DLM4G} on Factual Grounding (FGT) and Edit Sensitivity (ESR). The top block reports PLM baselines, the middle block reports diffusion baselines, and the bottom block reports \texttt{DLM4G} variants.}
\label{FGT}
\setlength{\tabcolsep}{6pt}
\renewcommand{\arraystretch}{1.15}
\resizebox{0.65\linewidth}{!}{
\begin{tabular}{lccccccc}
\toprule
\textbf{Model} & \textbf{Recall} & \textbf{F1} & \textbf{$|\mathcal{H}_\mathbf{S}|$} & \textbf{FGT@$\lambda$=0} & \textbf{FGT@$\lambda$=0.5} & \textbf{FGT@$\lambda$=1.0} & \textbf{ESR} \\
\midrule
GPT-2 (B)         & 0.60 & 0.65 & 2.95 & 0.65 & 0.59 & 0.53 & 0.46 \\
T5 (S)            & 0.58 & 0.62 & 3.10 & 0.62 & 0.56 & 0.50 & 0.42 \\
T5 (L)            & \underline{0.81} & \underline{0.83} & \underline{1.54} & \underline{0.83} & \underline{0.79} & \underline{0.75} & \underline{0.63} \\
\midrule
DiffuSeq          & 0.78 & 0.84 & 2.00 & 0.84 & 0.79 & 0.74 & 0.53 \\
SeqDiffuSeq       & \underline{0.80} & \underline{0.85} & 2.01 & \underline{0.85} & \underline{0.80} & \underline{0.74} & \underline{0.55} \\
\midrule
\rowcolor{white!70!yellow}
\texttt{DLM4G}-1.o & 0.80 & 0.79 & 2.03 & 0.79 & 0.74 & 0.70 & 0.60 \\
\rowcolor{white!70!yellow}
\texttt{DLM4G}-2.o & \textbf{0.82} & \textbf{0.86} & \textbf{1.08} & \textbf{0.86} & \textbf{0.83} & \textbf{0.80} & \textbf{0.68} \\
\rowcolor{white!70!gray}
\% Gain (vs.\ T5-L) & \textbf{+1.23\%} & \textbf{+3.61\%} & \textbf{29.8\%} & \textbf{+3.61\%} & \textbf{+5.16\%} & \textbf{+5.33\%} & \textbf{+7.9\%} \\
\rowcolor{white!70!gray}
\% Gain (vs.\ SeqDiffuSeq) & \textbf{+2.50\%} & \textbf{+1.18\%} & \textbf{46.3\%} & \textbf{+1.18\%} & \textbf{+3.75\%} & \textbf{+8.11\%} & \textbf{+23.6\%} \\
\bottomrule
\end{tabular}
}
\vspace{-4mm}
\end{table*}

\textbf{Summary (FGT, ESR)}: Table~\ref{FGT}, averaged across 100 edited examples, highlights two key trends.
\textit{First}, among the baselines, the best PLM (T5-Large) attains the lowest hallucination rate (1.54 entities/sequence) and strong scores (FGT@0 of 0.83, FGT@0.5 of 0.79 and ESR of 0.63), while the best diffusion baseline (SeqDiffuSeq) achieves higher FGT (0.85/0.80/0.74) but with more hallucinations (2.01) and lower ESR (0.55).
\textit{Second}, \texttt{DLM4G}-2.o consistently outperforms all baselines, improving over T5-Large in recall (0.82 vs. 0.81) while reducing hallucinations to a new low of 1.08 entities per sequence; consequently it yields the best overall grounding and edit sensitivity, improving FGT@0.5 by +5.16\% (vs.\ T5-L) and +3.75\% (vs.\ best diffusion) and ESR by +7.9\% and +23.6\%, respectively.

\vspace{-3mm}

\subsection{Ablation}

\noindent{\textbf{Graph-aware schedule:}
Table~\ref{ablation} compares the standard \textit{sqrt} schedule against our proposed Graph-aware schedule.
Applying Graph-aware noising to \textit{all} tokens improves overall generation quality (B: +0.03, M: +0.04) and increases semantic scores (B-F1: +0.03, MVE: +0.04),
while also improving grounding and edit response (FGT@0.5: +0.03, ESR: +0.05).
When we apply the Graph-aware schedule selectively to the graph-aligned tokens ($\mathcal{A}$) while keeping the standard schedule for others,
performance improves further, achieving the best B (\textbf{0.65}) and M (\textbf{0.79}) with strong B-F1 (\underline{0.96}) and MVE (\textbf{0.98}),
and the highest ESR (\textbf{0.68}) with FGT@0.5 matching the best (\textbf{0.83}).
This suggests the benefit comes not just from the schedule itself, but from differentiating the noise profile of factual entities from syntactic text.}

\textbf{Mapping Function}: We select the linear mapping primarily for its simplicity and ease of implementation. The impact of this design is visualized in Fig~\ref{sqrt}. As shown in Figure~\ref{sqrt}\textbf{L}, our adaptive schedule assigns a token-level noise schedule $\bar{\alpha}_t^i$ compared to the global \textit{sqrt} baseline. As seen by contrasting the loss profiles of syntactic tokens (Fig~\ref{sqrt}\textbf{M}) and factual tokens (Fig~\ref{sqrt}\textbf{R}), factual entities exhibit higher reconstruction difficulty but, under our graph-aware schedule, their loss evolves in a smooth, approximately linear over time. This stabilized, monotone difficulty profile keeps factual tokens informative throughout the trajectory and allows the denoiser to recover them more faithfully. While we also explored exponential, cosine, and polynomial mappings (Table~\ref{ablation}), we found they offered no performance benefit. Detailed comparisons of these alternatives are in App.~\ref{schedules}.

\noindent{\textbf{Inference efficiency (DDPM/DDIM):}}
Diffusion models incur iterative sampling costs at test time, so we quantify the quality--efficiency tradeoff under standard DDPM sampling and accelerated DDIM sampling.
On WikiOFGraph (batch size 50, single V100), \texttt{DLM4G}-2.o with standard DDPM sampling ($T=2000$ steps) takes 89s per batch and achieves 0.654 BLEU,
substantially faster than DiffuSeq (317s, 0.628 BLEU) while matching the latency of SeqDiffuSeq (89s, 0.616 BLEU).
To achieve faster sampling, we use DDIM with fewer steps $T'\!<\!2000$: at $T'=1000$ we obtain 0.551 BLEU (45s),
whereas at $T'=500$ performance drops to 0.365 BLEU (23s), below T5-Small (0.385; $<5$s), and at $T'=100$ we reach 0.312 BLEU (5s), remaining above GPT-2 (0.285; $<5$s).
We report the full wall-clock breakdown and additional metrics in App.~\ref{app:efficiency}. Overall, \texttt{DLM4G} is best operated at moderate-to-high step budgets, while very small $T'$ results in faster yet low quality output generation.

\vspace{-3mm}
\subsection{\texttt{DLM4G} for Molecule Captioning}
\texttt{DLM4G} has demonstrated strong performance in fluency (Section~\ref{results}) and factual grounding (Section~\ref{facts}). In this section we test its generalization to molecule captioning, a challenging G2S task from the scientific domain. 

\textbf{Dataset and Graph representation}: We use a subset of the M3-20M dataset \cite{guo2025m320mlargescalemultimodalmolecule} containing ~360,000 SMILES-description pairs, which we split 80/10/10 for training, validation, and testing. To process this data, we convert each SMILES string into a knowledge graph  \( \mathbf{G} \), where the molecule's atoms are treated as entities (nodes) and the chemical bonds between them are the relations (edges). This allows our model to interpret the molecule's structure. 

\textbf{Results (Molecule Captioning)}: First we analyze the scaling effect within the \texttt{DLM4G} variants. As shown in Fig \ref{molcap-overall}, the larger \texttt{DLM4G}-2.o (63M parameters) consistently outperforms the \texttt{DLM4G}-1.o version (50M). It achieves a +6.1\% improvement in BLEU, a +2.6\% gain in chrF++, and a significant +11.7\% increase in METEOR. This validates our scaling approach and establishes \texttt{DLM4G}-2.o as our best model.
More importantly, \texttt{DLM4G}-2.o achieves a new state-of-the-art result against all specialized baselines. The detailed analysis beside the table~\ref{tab:single_dataset_comparison} highlights the specific performance gains and the model's parameter efficiency. Refer App~\ref{App:mol} for details.

\noindent\textbf{Results (Table~\ref{tab:single_dataset_comparison}):}
Our \texttt{DLM4G}-2.o model outperforms all baselines across every metric.
It demonstrates strong performance on surface-level scores, achieving a BLEU of 0.567 (a +17.8\% gain over the best baseline),
and also leads on semantic metrics with a BERTScore-F1 of 0.843.
Crucially, it delivers these results while being 4$\times$ to 11$\times$ smaller than the baselines.

\vspace{-3mm}
\section{Conclusion and Limitations}\label{sec:limitations}

We presented \texttt{DLM4G}, a non-autoregressive diffusion framework for G2S generation. \texttt{DLM4G} targets two persistent limitations of fine-tuned autoregressive models: factual grounding and edit sensitivity. It learns a graph-aware noising schedule that preserves signal on graph-aligned (factual) tokens during training, while keeping unaligned syntactic tokens on the baseline schedule. At inference, when the alignment set is unavailable, \texttt{DLM4G} uses  cross-attention to linearly interpolate between the adaptive and baseline schedules. Across surface and embedding metrics, \texttt{DLM4G} improves over strong diffusion baselines; on task-grounded measures, it yields gains over comparably sized autoregressive models. Generalization to molecule captioning further supports the approach. Although \texttt{DLM4G} incurs diffusion-time sampling costs, resulting in quality-efficiency trade-off under reduced-step sampling. Performance also depends on alignment quality, and the current instantiation follows a dataset-defined KG linearization; thus it does not enforce permutation invariance and may be sensitive to alternative serializations. Future work will explore distillation, reduce reliance on explicit alignment, and incorporate structure-aware encoders that better preserve graph structure.

\section{Impact Statement}
This work implements graph-to-sequence generation by making diffusion-based text generation more faithful to an input graph and more sensitive to local graph edits (i.e., fewer hallucinated entities/relations and more localized text changes when the graph changes). If used carefully, this can help applications that verbalize structured knowledge (e.g., knowledge-graph summaries for QA and scientific captioning such as molecules) by producing more consistent, auditable descriptions. However, \texttt{DLM4G} can still generate incorrect or misleading text when the input graph is incomplete, the graph–text alignment is noisy, or the data distribution shifts, and it may reproduce biases present in training data; it can also pose privacy risks if graphs contain sensitive information. We therefore recommend using it with human review in high-stakes settings, showing the input graph alongside the output, and avoiding deployment on sensitive domains or data without appropriate safeguards.

\bibliography{example_paper}
\bibliographystyle{icml2026}

\newpage
\appendix
\onecolumn
\section{Appendix}
This section presents an in-depth discussion of the core components of the manuscript, including the principal mathematical derivations, template methods (zero-shot prompting and molecular captioning), the proposed algorithm pseudo-codes, and detailed implementation aspects. Additionally, the complete code is available here: \href{https://anonymous.4open.science/r/DLM4G-G2S-214B/README.md}{\textcolor{blue}{\texttt{CODE}}}
\section*{Contents}

\noindent\hyperref[sec:derivation_appendix]{\textbf{A.1} Derivation of Training Objective}%
\dotfill\pageref*{sec:derivation_appendix}\\

\noindent\hspace*{1.5em}\hyperref[app:forward_reverse]{A.1.1 Forward and Reverse Processes}%
\dotfill\pageref*{app:forward_reverse}\\
\noindent\hspace*{1.5em}\hyperref[app:vlb]{A.1.2 Variational Lower Bound (VLB)}%
\dotfill\pageref*{app:vlb}\\
\noindent\hspace*{1.5em}\hyperref[app:end-to-end]{A.1.3 Final end-to-end objective}%
\dotfill\pageref*{app:end-to-end}\\

\noindent\hyperref[App:related]{\textbf{A.2} Related Work and Background}%
\dotfill\pageref*{App:related}\\
\noindent\hspace*{1.5em}\hyperref[App:G2s_appedix]{A.2.1 Graph-to-Sequence (G2S)}%
\dotfill\pageref*{App:G2s_appedix}\\
\noindent\hspace*{1.5em}\hyperref[App:plm_appedix]{A.2.2 Pre-trained Language Models (PLMs) for Graphs}%
\dotfill\pageref*{App:plm_appedix}\\
\noindent\hspace*{1.5em}\hyperref[App:dlm_appedix]{A.2.3 Diffusion Models for Conditional Generation}%
\dotfill\pageref*{App:dlm_appedix}\\
\noindent\hspace*{1.5em}\hyperref[App:Mol]{A.2.4 Molecule Captioning}%
\dotfill\pageref*{App:Mol}\\

\noindent\hyperref[App:dataset_summary]{\textbf{A.3} Summary of Dataset}%
\dotfill\pageref*{App:dataset_summary}\\
\noindent\hspace*{1.5em}\hyperref[App:wiki]{A.3.1 WikiOFGraph}%
\dotfill\pageref*{App:wiki}\\
\noindent\hspace*{1.5em}\hyperref[App:genwiki]{A.3.2 GenWiki}%
\dotfill\pageref*{App:genwiki}\\
\noindent\hspace*{1.5em}\hyperref[App:tekgen]{A.3.3 TekGen}%
\dotfill\pageref*{App:tekgen}\\

\noindent\hyperref[app:alignment-analysis]{\textbf{A.4} Alignment Module}%
\dotfill\pageref*{app:alignment-analysis}\\
\noindent\hspace*{1.5em}\hyperref[app:align-setup]{A.4.1 Setup}%
\dotfill\pageref*{app:align-setup}\\
\noindent\hspace*{1.5em}\hyperref[app:align-overall]{A.4.2 Overall quality}%
\dotfill\pageref*{app:align-overall}\\
\noindent\hspace*{1.5em}\hyperref[app:align-alias]{A.4.3 Effect of alias-set size}%
\dotfill\pageref*{app:align-alias}\\
\noindent\hspace*{1.5em}\hyperref[app:align-example]{A.4.4 Example}%
\dotfill\pageref*{app:align-example}\\

\noindent\hyperref[gans]{\textbf{A.5} Graph-aware Noising Schedule Algorithm}%
\dotfill\pageref*{gans}\\
\noindent\hspace*{1.5em}\hyperref[app:ga-overview]{A.5.1 Overview}%
\dotfill\pageref*{app:ga-overview}\\
\noindent\hspace*{1.5em}\hyperref[app:ga-stage1]{A.5.2 Stage 1: Estimating token-wise denoising difficulty}%
\dotfill\pageref*{app:ga-stage1}\\
\noindent\hspace*{1.5em}\hyperref[app:ga-stage2]{A.5.3 Stage 2: Token-wise cummulative schedule}%
\dotfill\pageref*{app:ga-stage2}\\
\noindent\hspace*{1.5em}\hyperref[alg:graph_aware_training]{A.5.4 Algorithm 1: Graph-Aware Adaptive Noising}%
\dotfill\pageref*{alg:graph_aware_training}\\
\noindent\hyperref[app:toy_example]{\textbf{A.6} Worked Toy Example: Graph-Aware Adaptive Noising}%
\dotfill\pageref*{app:toy_example}\\
\noindent\hspace*{1.5em}\hyperref[app:toy-goal]{A.6.1 Goal}%
\dotfill\pageref*{app:toy-goal}\\
\noindent\hspace*{1.5em}\hyperref[app:toy-kg]{A.6.2 Toy KG and serialization}%
\dotfill\pageref*{app:toy-kg}\\
\noindent\hspace*{1.5em}\hyperref[app:toy-baseline]{A.6.3 Baseline schedule}%
\dotfill\pageref*{app:toy-baseline}\\
\noindent\hspace*{1.5em}\hyperref[app:toy-stage1]{A.6.4 Stage 1: Difficulty}%
\dotfill\pageref*{app:toy-stage1}\\
\noindent\hspace*{1.5em}\hyperref[app:toy-stage2]{A.6.5 Stage 2: Mapping}%
\dotfill\pageref*{app:toy-stage2}\\
\noindent\hspace*{1.5em}\hyperref[app:toy-reconstruct]{A.6.6 Reconstructing cummulative schedule}%
\dotfill\pageref*{app:toy-reconstruct}\\
\noindent\hspace*{1.5em}\hyperref[app:toy-snr]{A.6.7 SNR comparison}%
\dotfill\pageref*{app:toy-snr}\\
\noindent\hyperref[App:Train]{\textbf{A.7} Training Details and Hyperparameters}%
\dotfill\pageref*{App:Train}\\
\noindent\hyperref[app:G2S_sota]{\textbf{A.8} State-of-the-art G2S Baselines comparison}%
\dotfill\pageref*{app:G2S_sota}\\
\noindent\hyperref[schedules]{\textbf{A.9} Mapping Function Ablations}%
\dotfill\pageref*{schedules}\\
\noindent\hyperref[app:efficiency]{\textbf{A.10} Inference Efficiency Details}%
\dotfill\pageref*{app:efficiency}\\
\noindent\hyperref[App:FSZS]{\textbf{A.11} Zero-shot Prompting}%
\dotfill\pageref*{App:FSZS}\\
\noindent\hyperref[App:mol]{\textbf{A.12} Molecule Captioning}%
\dotfill\pageref*{App:mol}\\

\subsection{Derivation of the Training Objective}
\label{sec:derivation_appendix}

\texttt{DLM4G} builds on the standard diffusion framework, which trades the flexibility of expressive generative models (e.g., GANs, VAEs, flow models) for the tractability of likelihood-based training in a continuous latent space $\mathbf{z}$. The overall goal is to minimize the negative log-likelihood
\begin{equation}
    \mathbb{E}_{\mathbf{z}_0,\mathbf{c}}\!\bigl[-\log p_\theta(\mathbf{z}_0 \mid \mathbf{c})\bigr],
\end{equation}
which is upper-bounded by the Variational Lower Bound (VLB).

\subsubsection{Forward and Reverse Processes}\label{app:forward_reverse}

The forward Markov chain is defined as $ q(\mathbf{z}_{1:T} \mid \mathbf{z}_0)
    = \prod_{t=1}^T q(\mathbf{z}_t \mid \mathbf{z}_{t-1}),$
where each transition is Gaussian:
\begin{equation}
    q(\mathbf{z}_t \mid \mathbf{z}_{t-1})
    = \mathcal{N}\!\Bigl(
        \mathbf{z}_t \,\big|\,
        \sqrt{1-\beta_t}\,\mathbf{z}_{t-1},\,
        \beta_t\,\mathbf{I}
    \Bigr).
\end{equation}
Let $\alpha_t = 1-\beta_t$ and $\bar{\alpha}_t = \prod_{i=1}^t \alpha_i$.  
By induction, the marginal at time $t$ satisfies:
\begin{equation}
    \mathbf{z}_t
    = \sqrt{\bar{\alpha}_t}\,\mathbf{z}_0
    + \sqrt{1-\bar{\alpha}_t}\,\boldsymbol{\epsilon},
    \qquad
    \boldsymbol{\epsilon}\sim\mathcal{N}(\mathbf{0},\mathbf{I}),
\end{equation}
so that $ q(\mathbf{z}_t \mid \mathbf{z}_0)
    = \mathcal{N}\!\Bigl(
        \sqrt{\bar{\alpha}_t}\,\mathbf{z}_0,\,
        (1-\bar{\alpha}_t)\mathbf{I}
    \Bigr).$
We used the \textit{sqrt} schedule as the baseline schedule used in DiffusionLM \cite{li2022diffusionlmimprovescontrollabletext}, namely \(\bar\alpha_t=1-\sqrt{t/T+s}\) with small \(s>0\).
The reverse denoising process then learns
\begin{equation}
p_\theta(\mathbf{z}_{0:T})
= p(\mathbf{z}_T)\prod_{t=1}^T
p_\theta(\mathbf{z}_{t-1}\mid \mathbf{z}_t),
\quad
p_\theta(\mathbf{z}_{t-1}\mid \mathbf{z}_t)
=\mathcal{N}\!\bigl(\boldsymbol{\mu}_\theta(\mathbf{z}_t,t),\,\boldsymbol{\sigma}_\theta^2(\mathbf{z}_t,t)\bigr).
\end{equation}
Applying Bayes’ rule to the forward transitions yields the exact posterior mean
\begin{equation}
\boldsymbol{\mu}_t(\mathbf{z}_t,\mathbf{z}_0)
= \frac{\sqrt{\alpha_t}(1-\bar\alpha_{t-1})}{1-\bar\alpha_t}\,\mathbf{z}_t
  + \frac{\sqrt{\bar\alpha_{t-1}}\,\beta_t}{1-\bar\alpha_t}\,\mathbf{z}_0,
\end{equation}
whose coefficients we denote by \(\mathcal{U}\) and \(\mathcal{E}\).  \texttt{DLM4G}’s training objective is then to match the network’s predicted \(\boldsymbol{\mu}_\theta,\boldsymbol{\sigma}_\theta\) to these posterior quantities via a simple noise‐prediction loss. 
We optimize the negative log‐likelihood by upper‐bounding it with the variational lower bound
\begin{equation}
\mathbb{E}\bigl[-\log p_\theta(x_0)\bigr]
\;\le\;
\mathcal{L}_{\mathrm{vlb}}
\;=\;
\sum_{t=0}^T \mathcal{L}_t.
\end{equation}

\subsubsection{Variational Lower Bound (VLB)}\label{app:vlb}

Following Sohl‐Dickstein et al.\cite{pmlr-v37-sohl-dickstein15}, for conditional generation the VLB decomposes into:
\begin{equation}
    \mathcal{L}_{\mathrm{vlb}}
    =
    \mathbb{E}_{q(\mathbf{z}_{1:T}\mid\mathbf{z}_0)}
    \Biggl[
        \underbrace{\log\frac{q(\mathbf{z}_T\mid \mathbf{z}_0)}{p(\mathbf{z}_T)}}_{\mathcal{L}_T}
        + \sum_{t=2}^T
        \underbrace{\log \frac{
            q(\mathbf{z}_{t-1} \mid \mathbf{z}_t,\mathbf{z}_0)
        }{
            p_\theta(\mathbf{z}_{t-1}\mid \mathbf{z}_t,\mathbf{c})
        }}_{\mathcal{L}_t}
        - \underbrace{\log p_\theta(\mathbf{z}_0 \mid \mathbf{z}_1,\mathbf{c})}_{\mathcal{L}_0}
    \Biggr],
\end{equation}
where each $\mathcal{L}_t$ is a KL divergence between Gaussians.  
The true posterior mean (via Bayes' rule) is:
\begin{equation}
    \boldsymbol{\mu}_t(\mathbf{z}_t,\mathbf{z}_0)
    =
    \underbrace{
        \frac{\sqrt{\alpha_t}(1-\bar{\alpha}_{t-1})}
             {1-\bar{\alpha}_t}
    }_{\mathcal{U}}
    \mathbf{z}_t
    +
    \underbrace{
        \frac{\sqrt{\bar{\alpha}_{t-1}}\beta_t}
             {1-\bar{\alpha}_t}
    }_{\mathcal{E}}
    \mathbf{z}_0,
\end{equation}
with covariance $\boldsymbol{\Sigma}_q = \tilde{\beta}_t \mathbf{I}$, $\tilde{\beta}_t =
    \frac{1-\bar{\alpha}_{t-1}}{1-\bar{\alpha}_t}\beta_t.$
In the standard simplification, the model's covariance is fixed to match the true posterior covariance $\boldsymbol{\Sigma}_\theta = \boldsymbol{\Sigma}_q$,  
the KL collapses to a weighted MSE:
\begin{equation}
    \mathcal{L}_t
    = \frac{1}{2}
        \bigl\|
            \boldsymbol{\mu}_t
            - \boldsymbol{\mu}_\theta
        \bigr\|_{\boldsymbol{\Sigma}_q^{-1}}^2
    \;\propto\;
    \mathbb{E}\!
    \left[
        \bigl\|
            \mathbf{z}_0
            - \mathcal{M}_\theta(\mathbf{z}_t,t,\mathbf{c})
        \bigr\|^2
    \right].
\end{equation}
Thus, for $2\le t\le T$, $ \mathcal{L}_t
    \rightarrow
    \|\mathbf{z}_0
    - \mathcal{M}_\theta(\mathbf{z}_t,t,\mathbf{c})\|^2.$
The final KL encourages $\mathbf{z}_T$ to match the unit Gaussian prior:
\begin{equation}
    \mathcal{L}_T
    = \mathrm{KL}\bigl(q(\mathbf{z}_T\mid\mathbf{z}_0)\,\|\,p(\mathbf{z}_T)\bigr)
    \;\propto\;
    \bigl\|\boldsymbol{\mu}(\mathbf{z}_T)\bigr\|^2,
\end{equation}
a constant w.r.t.\ $\theta$.
The discrete target $\mathbf{S}$ (sequence) is encoded into a continuous embedding $g_\Phi(\mathbf{S})$.  
The final term in VLB is $\mathcal{L}_0 = -\log p_\theta(\mathbf{z}_0\mid\mathbf{z}_1, \mathbf{c})$. We need to integrate the discrete data $\mathbf{S}$ into this continuous likelihood term.
We use the law of total probability to express the continuous likelihood $p_\theta(\mathbf{z}_0\mid\mathbf{z}_1, \mathbf{c})$ by marginalizing over all possible discrete tokens in the target sequence $\mathbf{S} = \{s_1, s_2 \cdots, s_N\}$: 
\begin{equation}
    p_\theta(\mathbf{z}_0 \mid \mathbf{z}_1, \mathbf{c}) = \sum_{\mathbf{S}} p_\theta(\mathbf{z}_0, \mathbf{S} \mid \mathbf{z}_1, \mathbf{c})
\end{equation}
We then apply the product rule to the joint probability:
\begin{equation}
    p_\theta(\mathbf{z}_0, \mathbf{S} \mid \mathbf{z}_1, \mathbf{c}) = p_\theta(\mathbf{z}_0 \mid \mathbf{S}, \mathbf{z}_1, \mathbf{c}) \cdot p_\theta(\mathbf{S} \mid \mathbf{z}_1, \mathbf{c})
\end{equation}
For training, we are interested in the specific ground-truth sequence $\mathbf{S}$. When we evaluate $\mathcal{L}_0$ during training, we consider only the term where $\mathbf{S}$ is the ground-truth sequence:
\begin{equation}
    \mathcal{L}_0 \approx -\log p_\theta(\mathbf{z}_0, \mathbf{S} \mid \mathbf{z}_1, \mathbf{c}) = -\log \left[ p_\theta(\mathbf{z}_0 \mid \mathbf{S}, \mathbf{z}_1, \mathbf{c}) \cdot p_\theta(\mathbf{S} \mid \mathbf{z}_1, \mathbf{c}) \right]
\end{equation}
The core approximation simplifies the dependency graph by asserting that the discrete data $\mathbf{S}$ is generated only from the clean latent $\mathbf{z}_0$, and is independent of $\mathbf{z}_1$ and $\mathbf{c}$ given $\mathbf{z}_0$.
\begin{equation}
    \mathbf{S} \perp (\mathbf{z}_1, \mathbf{c}) \mid \mathbf{z}_0
\end{equation}
This allows us to replace the discrete conditional likelihood with the separate rounding network $\tilde{p}_\Phi(\mathbf{S} \mid \mathbf{z}_0)$:
$p_\theta(\mathbf{S} \mid \mathbf{z}_1, \mathbf{c}) \approx \tilde{p}_\Phi(\mathbf{S} \mid \mathbf{z}_0)$.
Substituting this back into the likelihood decomposition:
\begin{equation}
    p_\theta(\mathbf{z}_0, \mathbf{S} \mid \mathbf{z}_1, \mathbf{c}) \approx p_{\mathrm{cont}}(\mathbf{z}_0 \mid \mathbf{S}, \mathbf{z}_1, \mathbf{c}) \cdot \tilde{p}_\Phi(\mathbf{S} \mid \mathbf{z}_0)
\end{equation}
Taking the negative logarithm of the approximation gives the two desired terms:

$$\mathcal{L}_0 \approx -\log p_{\mathrm{cont}}(\mathbf{z}_0 \mid \mathbf{S}, \mathbf{z}_1, \mathbf{c}) - \log \tilde{p}_\Phi(\mathbf{S} \mid \mathbf{z}_0)$$

This split yields the two components used in the final training objective:
\begin{enumerate}
    \item Consistency Term ($\mathcal{L}_{\mathrm{Cons}}$): The first term is the negative log-likelihood of the continuous latent, which is minimized via the MSE loss on the means:
    $-\log p_{\mathrm{cont}}(\mathbf{z}_0 \mid \mathbf{S}, \mathbf{z}_1, \mathbf{c}) \rightarrow \mathcal{L}_{\mathrm{Consistency}} = \bigl\| g_\Phi(\mathbf{S}) - \mathcal{M}_\theta(\mathbf{z}_1,1,\mathbf{c}) \bigr\|^2$.
    \item Rounding Term ($\mathcal{L}_{\text{Round}}$): This second term is the dedicated loss for the discrete data likelihood:
    $\mathcal{L}_{\text{Round}} = -\log \tilde{p}_\Phi(\mathbf{S} \mid \mathbf{z}_0)$
\end{enumerate}

\subsubsection{Final End-to-End Objective}\label{app:end-to-end}

Combining all components:
\begin{align}
    \mathcal{L}_{\mathrm{vlb}}
    \propto
    \sum_{t=2}^T
    \underbrace{
        \bigl\|
            \mathbf{z}_0
            - \mathcal{M}_\theta(\mathbf{z}_t,t,\mathbf{c})
        \bigr\|^2
    }_{\text{Denoising}}
    +
    \underbrace{
        \bigl\|
            g_\Phi(\mathbf{S})
            - \mathcal{M}_\theta(\mathbf{z}_1,1,\mathbf{c})
        \bigr\|^2
    }_{\text{Consistency}}
    -
    \underbrace{
        \log \tilde{p}_\Phi(\mathbf{S}\mid\mathbf{z}_0)
    }_{\text{Rounding}}.
\end{align}
Dropping constant terms, the simplified end-to-end training loss is:
\begin{equation}
\resizebox{0.9\linewidth}{!}{
    $ \displaystyle
    \mathcal{L}_{\text{e2e-simple}}(\mathbf{S}) = \mathbb{E}_{q} 
    \Bigl[ \sum_{t=2}^{T} \underbrace{\|\mathcal{M}_\theta(\mathbf{z}_t, t, \mathbf{G}) - \mathbf{z}_0 \|^2}_{\text{Denoising}} + \underbrace{\|\,g_\Phi(\mathbf{S}) - \mathcal{M}_\theta(\mathbf{z}_1, 1, \mathbf{G})\|^2}_{\text{Consistency}} - \underbrace{\log \tilde{p}_\Phi(\mathbf{S} \mid \mathbf{z}_0)}_{\text{Rounding}} \Bigr]
    $
}
\end{equation}

\subsection{Related Work and Background}\label{App:related}
\textbf{Graph-to-Sequence Learning}\label{App:G2s_appedix}: G2S has evolved through three stages: (i) template-based systems that verbalised graph predicates but were brittle for complex inputs \cite{vejvar-fujimoto-2023-aspiro}; (ii) neural encoder–decoder models that learned graph embeddings, improving structural generalisation yet struggling with long-range dependencies \cite{iso-etal-2019-learning}; and (iii) fine-tuned transformers, now dominant, offering superior fluency and factuality with minimal task-specific design \cite{han-shareghi-2022-self}. This trajectory frames the current G2S landscape and motivates subsequent approaches.

\textbf{PLMs for Graphs}\label{App:plm_appedix}: Leveraging LLMs for graph verbalisation involves four challenges: (i) \emph{alignment} of graph elements to words \cite{luo2024enhancegraphalignmentlarge, zhu2025llmgnngraphvocabulary}, (ii) \emph{position} encoding under permutation invariance \cite{10.5555/3692070.3692236, huang2024on, perozzi2024letgraphtalkingencoding}, (iii) \emph{multi-level semantics} across nodes, edges, and subgraphs \cite{wang2024mgsamultigranularitygraphstructure}, and (iv) \emph{context} retention over long spans \cite{ding-etal-2025-msg, wang-etal-2024-bridging-local}. These define a taxonomy from Graph-to-Sequence (G2S) to Graph-to-Token (G2T) methods. Current KG-to-text models employ positional encodings, structural prompts, and multi-granularity attention \cite{luo2024enhancegraphalignmentlarge, zhu2025llmgnngraphvocabulary, wang2024mgsamultigranularitygraphstructure}, reducing factual omissions but still limited by left-to-right decoding and weak global planning \cite{wei2022emergent, lin-etal-2021-limitations}. Diffusion LMs, with iterative denoising, could address these issues, though they remain unexplored for KG-to-text generation \cite{10.24963/ijcai.2023/750}.

\textbf{Diffusion Models for Conditional Generation}\label{App:dlm_appedix}: Conditional diffusion guides denoising with an input sequence encoding, extending conditional-VAE ideas \cite{zhao-etal-2017-learning}. Early text models (Diffusion-LM \cite{li2022diffusionlmimprovescontrollabletext}, Analog Bits \cite{chen2023analog}) imposed weak conditioning via classifiers or plug-in controls, while DIFFUSEQ \cite{gong2023diffuseqsequencesequencetext, yuan-etal-2024-text} enabled true sequence-to-sequence conditioning in continuous space. Related frameworks also target time-series (CSDI \cite{10.5555/3540261.3542161}) and speech (WaveGrad \cite{chen2021wavegrad}). Distinct from prior G2S and diffusion-LM work, \texttt{DLM4G} integrates classifier-free diffusion with explicit KG conditioning, treating the graph itself as the control variable. This eliminates exposure bias and supports global planning, yielding more coherent KG verbalisation.

\textbf{Molecule Captioning:}\label{App:Mol}  
Most prior works adapt either AR or NAR generation for molecular descriptions, but these methods often inherit exposure bias (AR) or strong independence assumptions (NAR) \cite{LIU2024108073}. Diffusion-based approaches, while promising for text generation, have not been systematically applied to graph-to-sequence captioning. To clarify the conceptual distinctions, Table~\ref{tab:comparison} summarizes the characteristics of major generation paradigms and highlights how \texttt{DLM4G} differs. In particular, our method introduces a \textit{graph-guided refinement process} with \textit{graph-aware noising}, enabling both factual grounding and graph edits during caption generation, a capability absent in existing paradigms.

\begin{table*}[h]
\centering
\caption{Comparison of \texttt{DLM4G} with existing paradigms (FG: Factual Grounding; GE: Graph Edits).}
\renewcommand{\arraystretch}{1.4}
\resizebox{\textwidth}{!}{%
\begin{tabular}{@{}l|l|c|l|l@{}}
\toprule
\textbf{Model Family} & \textbf{Output Generation Paradigm} & \textbf{Noising Schedule} & \textbf{Mechanism for FG / GE} & \textbf{Molecule Captioning} \\ \midrule
\textbf{Autoregressive (AR)} & Sequential, left-to-right token prediction &  No Diffusion & Implicit (data-driven/ graph prompts ) & Standard G2S application \\
 & \textit{(Exposure bias, local optima}, \textit{e.g., BART, T5)} &  &  & \\
\midrule
\textbf{Non-Autoregressive (NAR)} & Parallel, independent token prediction & No Diffusion  & Implicit (sequence-level objective; no graph-aware bias) & Standard G2S application \\
 & \textit{(Conditional independence assumption)} \textit{(e.g., Mask-Predict)} &  &  & \\
\midrule
\textbf{Standard Diffusion LMs} & Iterative, parallel refinement from noise & Uniform, Isotropic & Implicit (standard diffusion schedule) & Unexplored for G2S; \\
 & \textit{Advantage: Mitigates exposure bias}\textit{(e.g., DiffuSeq)} &  &  & applied to S2G (generation) \\
\midrule
\rowcolor{yellow!30} 
\textbf{\texttt{DLM4G} (Ours)} & \textbf{Iterative, graph-guided refinement} & \textbf{Graph-aware noising} & \textbf{Explicit (graph-aware schedule over $\mathcal{A}$)} & \textbf{Novel G2S application} \\
\rowcolor{yellow!30} 
& \textit{(Global planning + factual grounding)} & \textit{(Preserves entity, relations)} &  & \textit{(Graph }$\rightarrow$\textit{ Sequence task)} \\
\bottomrule
\end{tabular}%
}
\label{tab:comparison}
\end{table*}

\subsection{Summary of Datasets}\label{App:dataset_summary}
\begin{table}[h!]
  \centering
  \caption{Training set statistics for comparative analysis. \textit{\# triplet (m/M/avg)} indicates the minimum, maximum, and average number of triplets per sample.}
  \label{tab:dataset_stats}
  \begin{tabular}{lrrrc}
    \toprule
    \textbf{Dataset} & \textbf{\# samples} & \textbf{\# unique predicate} & \textbf{\# unique entity} & \textbf{\# triplet (m/M/avg)} \\
    \midrule
    WikiOFGraph  & 5.85M & 140,733 & 8.2M & 1/173/3.62 \\
    GenWiki & 680K & 287 & 86.6K & 1/10/2.64 \\
    TekGen & 6.31M & 50,861 & 4.3M & 1/54/1.73 \\
    
    \bottomrule
  \end{tabular}
\end{table}

\textbf{WikiOFGraph}\label{App:wiki}: We use the WikiOFGraph dataset as described in \cite{kim2024ontology}. This dataset comprises approximately 5.85 million graph–text pairs extracted from general‐domain English Wikipedia articles. Each graph is represented as a set of RDF‐style triples, automatically mined and refined via large‐language‐model prompting. For example, the triple \texttt{<Alan Turing, birthPlace, London>} corresponds to the sentence “Alan Turing was born in London.”

\textbf{GenWiki}\label{App:genwiki}: We use the “fine” split of GenWiki \cite{jin-etal-2020-genwiki}, which contains 680 K graph–text pairs; we reserve 10 \% of these for evaluation. The dataset covers 287 distinct predicates, with an average of $2.64\pm1.72$ triples per graph and an average text length of $26.05\pm10.99$ tokens. For instance, the graph \{\,(Google, founder, Larry Page), (Google, founder, Sergey Brin)\} maps to the sentence “Google was founded by Larry Page and Sergey Brin.”

\textbf{TekGen}\label{App:tekgen}: We adopt the TekGen dataset as released in \cite{mousavi-etal-2024-construction}, containing roughly 6.3 M aligned Wikidata triple–sentence pairs drawn from Wikipedia. It spans about 50.8 K distinct predicates and is provided in separate train/validation/test TSV files (each line a JSON object). An exemplar entry is:  
  \textit{\{"subject":"The Lion King","predicate":"director","object":"Roger Allers","text":"The Lion King is an animated musical drama film directed by Roger Allers and Rob Minkoff."\}}

\subsection{Alignment Module}
\label{app:alignment-analysis}

\begin{figure}[h!]
    \centering
    \includegraphics[width=\linewidth]{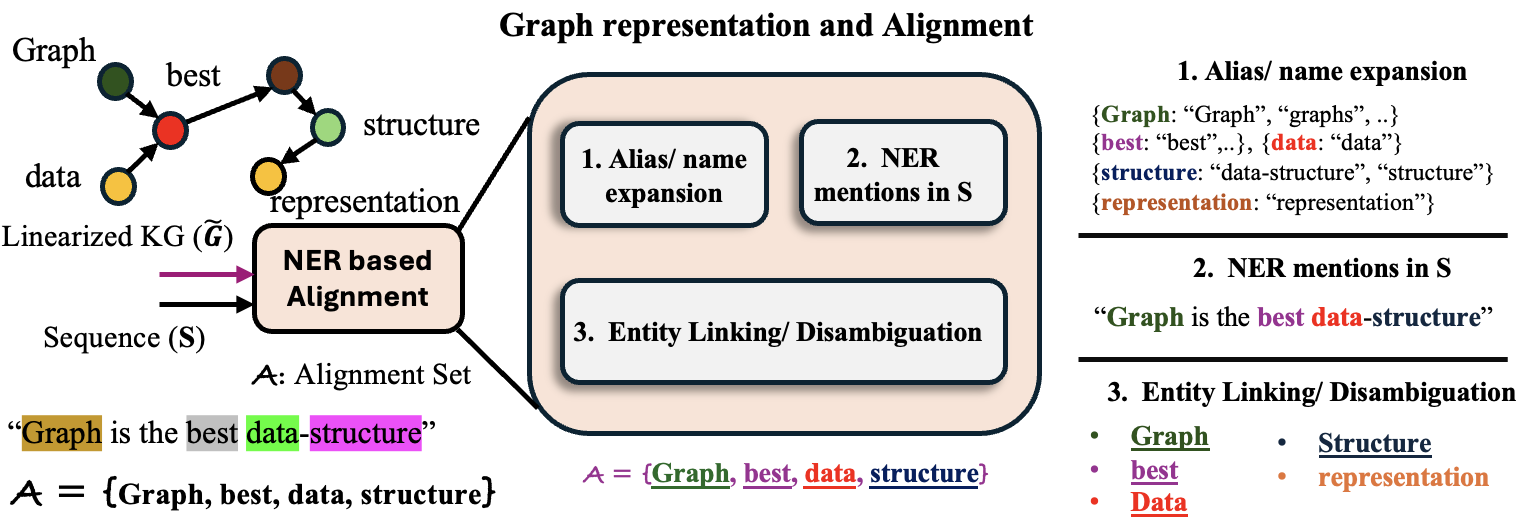}
    \caption{NER-based graph--sequence alignment. 
    Given a linearized graph $\mathbf{G}$ and target sequence $\mathbf{S}$, 
    the module (1) expands aliases for each KG node, (2) detects mentions in $\mathbf{S}$,
    and (3) links/disambiguates them to obtain the alignment set 
    $\mathcal{A}$ of graph-grounded tokens.}
    \label{fig:alignment-module}
\end{figure}

\textbf{Setup.}\label{app:align-setup}
For each graph--sequence pair $(\mathbf{G}, \mathbf{S})$, the alignment module
outputs a set $\mathcal{A}$ of token spans in $\mathbf{S}$ that are linked to
entities or relations in $\mathbf{G}$ (see Fig.~\ref{fig:alignment-module} 
and Sec.~\ref{alignment}).
To quantify the quality of this mapping, we manually annotate a subset of
WikiOFGraph dev examples (100 examples) with gold alignments $\mathcal{A}^{\star}$ and evaluate
the module at the span level: a prediction is correct (TP) if the span overlaps
a gold mention and is linked to the same KG node; other predicted spans are
counted as FP, and unmatched gold spans as FN.
We report precision, recall, and F1 over $(\text{span},\text{KG node})$ pairs,
as well as token and KG-node coverage.

\textbf{Overall quality.}\label{app:align-overall}
Table~\ref{tab:alignment-quality} summarizes the quality of the alignment
module for our default configuration (maximum of $k{=}5$ aliases per KG node).
Specifically, \textbf{Token coverage} measures the percentage of tokens in the target sequence $\mathbf{S}$ that are part of an aligned span, while \textbf{KG-node coverage} measures the percentage of entities and relations in $\mathbf{G}$ that successfully link to $\mathbf{S}$.
The module achieves high precision while covering a substantial fraction of
graph-grounded tokens and KG nodes, which is sufficient to anchor the 
graph-aware noising schedule.

\begin{table}[h!]
\centering
\caption{Alignment quality on the WikiOFGraph
(dev), with $k{=}5$ aliases per KG node. The average magnitude of the alignment set ($|\mathcal{A}|$) represents the size of alignment set per example.
Token coverage is the percentage of target tokens that belong to some aligned
span; KG-node coverage is the percentage of nodes in $\mathbf{G}$
with at least one aligned mention in $\mathbf{S}$.}
\label{tab:alignment-quality}
\begin{tabular}{lcccccc}
\toprule
\textbf{Setting} & \textbf{Prec.} & \textbf{Rec.} & \textbf{F1} & \textbf{Token cov. (\%)} & \textbf{KG-node cov. (\%)} & \textbf{$|\mathcal{A}|$} \\
\midrule
Aliases  ($k{=}5$) & 0.90 & 0.78 & 0.83 & 24.1 & 86.3 & 7.4 \\ 
\bottomrule
\end{tabular}
\end{table}

\textbf{Effect of alias-set size.}\label{app:align-alias}
The size of the alias set controls the effective size of $\mathcal{A}$:
larger $k$ exposes more surface forms and increases recall and coverage, but
can introduce additional ambiguity and harm precision.
We vary $k \in \{2,3,4,5\}$ and re-evaluate the module on the same annotated
subset, as well as the downstream performance of \texttt{DLM4G} on
WikiOFGraph (Table~\ref{tab:alias-ablation}).
We observe a smooth precision--recall trade-off as $k$ increases; the default
$k{=}5$ offers a good balance, yielding the best BLEU score.

\begin{table}[h!]
\centering
\caption{Alias-budget ablation on WikiOFGraph. 
Increasing the maximum number of aliases $k$ per KG node improves recall and
coverage but slightly reduces precision, resulting in a modest but consistent
gain in downstream performance.}
\label{tab:alias-ablation}
\begin{tabular}{cccccccc}
\toprule
\textbf{$k$ (aliases)}
& \textbf{Prec.} & \textbf{Rec.} & \textbf{F1}
& \textbf{$|\mathcal{A}|$}
& \textbf{Token cov. (\%)} & \textbf{KG-node cov. (\%)} & \textbf{BLEU} \\
\midrule
2 & 0.94 & 0.70 & 0.80 & 3.9 & 17.3 & 75.1 & 0.603 \\
3 & 0.92 & 0.75 & 0.83 & 4.3 & 20.5 & 80.4 & 0.609 \\
4 & 0.91 & 0.77 & 0.84 & 6.7 & 22.8 & 84.0 & 0.624 \\
5 & 0.90 & 0.78 & 0.83 & 7.4 & 24.1 & 86.3 & 0.651 \\
\bottomrule
\end{tabular}
\end{table}
\textbf{Example.}\label{app:align-example}
To illustrate the alignment process, consider the following graph--sequence
pair:\\
Serialized KG ($\mathbf{G}$): \texttt{$\langle$[HEAD] USA [REL] hosted [TAIL] 1994\_FIFA\_World\_Cup$\rangle$ [SEP] $\langle$[HEAD] USA [REL] capital [TAIL] Washington\_D.C.$\rangle$ [SEP] $\langle$[HEAD] 1994\_FIFA\_World\_Cup [REL] top\_scorer [TAIL] Hristo\_Stoichkov$\rangle$}. \\ Corresponding sequence ($\mathbf{S}$): \textit{``The United States hosted the 1994 FIFA World Cup; its capital is Washington, D.C., and the tournament's top scorer was Hristo Stoichkov''.}\\
\emph{(1) Alias expansion.}  
From the KG we construct an alias dictionary, e.g.,
\begin{align*}
\text{USA: } & \{\text{``USA''}, \text{``U.S.''}, \text{``United States''}, 
\text{``United States of America''}, \dots\},\\
\text{1994\_FIFA\_World\_Cup: } & 
\{\text{``1994 FIFA World Cup''}, \text{``1994 World Cup''}, \dots\},\\
\text{Washington\_D.C.: } & 
\{\text{``Washington, D.C.''}, \text{``Washington DC''}, \dots\},\\
\text{Hristo\_Stoichkov: } & 
\{\text{``Hristo Stoichkov''}, \text{``Stoichkov''}\}.
\end{align*}
\noindent
\emph{(2) NER mentions in $\mathbf{S}$.}  
A NER detector identifies mentions such as
“United States”, “1994 FIFA World Cup”, “Washington, D.C.”, and “Hristo
Stoichkov” in the sequence.\\
\noindent
\emph{(3) Entity linking / disambiguation.}  
Each mention is matched against the alias dictionary and, if multiple
candidates exist, disambiguated using local context similarity to KG node
descriptions.
For this example the module recovers the alignment set $\mathcal{A}$ as:\\
$\mathcal{A} = \{(\text{``United States''}, \text{USA}),\;
(\text{``1994 FIFA World Cup''}, \text{1994\_FIFA\_World\_Cup}),\;\\
(\text{``Washington, D.C.''}, \text{Washington\_D.C.}),\;
(\text{``Hristo Stoichkov''}, \text{Hristo\_Stoichkov}),\\
(\text{``hosted''}, \text{hosted}),\;
(\text{``capital''}, \text{capital}),\;
(\text{``top scorer''}, \text{top\_scorer})\}$$\},$\\
which corresponds to seven true-positive links between $\mathbf{S}$ and
$\mathbf{G}$.
This alignment set is then used to derive token-wise difficulty profiles and
the graph-aware noising schedule described in Sec.~\ref{sec:graph_aware_noising}.\\
\textbf{Details.} For this specific instance, the sequence $\mathbf{S}$ contains 32 tokens. The $\mathbf{G}$ contains 7 unique factual elements (4 entities and 3 relations). The alignment set $\mathcal{A}$ has a magnitude $|\mathcal{A}|=\mathbf{7.0}$, aligning all 7 elements (100.0\% KG-node coverage). The results are reported in Table~\ref{tab:align-example-metrics}.
\begin{table}[h!]
\centering
\caption{Alignment Metrics for the example $(\mathbf{G},\mathbf{S})$ pair
.}
\label{tab:align-example-metrics}
\begin{tabular}{cccccccc}
\toprule
\textbf{$k$ (aliases)}
& \textbf{Prec.} & \textbf{Rec.} & \textbf{F1}
& \textbf{$|\mathcal{A}|$}
& \textbf{Token cov. (\%)} & \textbf{KG-node cov. (\%)} & \textbf{BLEU} \\
\midrule
5 & 1.00 & 1.00 & 1.00 & 7.0 & 31.2 & 100.0 & 0.773 \\
\bottomrule
\end{tabular}
\end{table}

\subsection{Graph-aware Noising Schedule Algorithm}\label{gans}
\noindent\textbf{Overview.}\label{app:ga-overview}
Algorithm~\ref{alg:graph_aware_training} implements the two-stage procedure described in
Sec.~\ref{sec:graph_aware_noising} for constructing \emph{token-wise} diffusion schedules on aligned positions
$i\in\mathcal{A}$, while leaving unaligned tokens $i\notin\mathcal{A}$ on the baseline schedule.
Concretely, we maintain per-token per-step coefficients $\{\alpha_{t,i}\}_{t=1}^T$ and reconstruct the
cumulative schedule via $\bar{\alpha}^{\,i}_t=\prod_{k=1}^t \alpha_{k,i}$ (with $\bar{\alpha}^{\,i}_0=1$).
The procedure is updated every $K_{\mathrm{up}}$ training steps to amortize cost.

\noindent\textbf{Stage 1 (difficulty estimation).}\label{app:ga-stage1}
For each aligned token $i\in\mathcal{A}$, we estimate the denoising difficulty curve
$\{\ell_t^{\,i}\}_{t=1}^T$ using Eq.~\ref{eq:token_difficulty}. To stabilize the mapping and reduce variance,
we aggregate these losses over non-overlapping windows $W_m$ of length $K_{\mathrm{win}}$ to obtain
window-level difficulties $\tilde{\ell}_m^{\,i}$, and normalize them using the global window extrema
$\ell_m^{\min},\ell_m^{\max}$ across $i\in\mathcal{A}$.

\noindent\textbf{Stage 2 (monotone loss-to-noise mapping).}\label{app:ga-stage2}
Given $\tilde{\ell}_m^{\,i}$, we assign a window-level coefficient $\tilde{\alpha}_{m,i}$ by mapping
difficulty to signal strength using the monotone mapping $\Psi_m(\cdot)$ (Eq.~\ref{eq:window_mapping}),
anchored between the baseline coefficients $\alpha_{t_m}^{\mathrm{base}}$ and $\alpha_{t_{m-1}}^{\mathrm{base}}$.
This enforces the intended behavior \emph{higher difficulty $\Rightarrow$ less noising}
(i.e., larger $\alpha$), while clipping to $[0,1]$ ensures validity.
Finally, the token-wise cumulative schedules are reconstructed from $\{\alpha_{t,i}\}$ and returned for use in
the forward diffusion process during training.

\begin{algorithm}[h!]
\caption{Graph-Aware Adaptive Noising}
\label{alg:graph_aware_training}
\begin{algorithmic}[1]
\REQUIRE Baseline cumulative schedule $\{\bar{\alpha}_t\}_{t=1}^T$,
alignment set $\mathcal{A}$,
update interval $K_{\mathrm{up}}$,
window size $K_{\mathrm{win}}$,
constant $\tau>0$,
minimum coefficient $\alpha_{\min}\in(0,1)$
\ENSURE Current token-wise cumulative schedules $\{\bar{\alpha}^{\,i}_{t}\}_{t=1}^T$ for all $i\in\{1,\dots,N\}$

\STATE Compute baseline coefficients
$\alpha_t^{\mathrm{base}} \leftarrow \bar{\alpha}_t/\bar{\alpha}_{t-1}$ (with $\bar{\alpha}_0=1$).

\STATE Initialize current schedules:
$\bar{\alpha}^{\,i}_{t}\leftarrow\bar{\alpha}_t$ for all $i,t$.

\IF{$\texttt{train\_step} \bmod K_{\mathrm{up}} = 0$}
  \STATE Let $M \leftarrow \left\lceil \frac{T}{K_{\mathrm{win}}}\right\rceil$ and
  $W_m \leftarrow \{(m-1)K_{\mathrm{win}}+1,\dots,\min(mK_{\mathrm{win}},T)\}$.

  \FOR{$i \in \mathcal{A}$}
    \STATE \textbf{Stage 1:} Estimate $\{\ell_t^{\,i}\}_{t=1}^T$ via Eq.~\ref{eq:token_difficulty}.
    \STATE Compute $\tilde{\ell}_m^{\,i}\leftarrow \frac{1}{|W_m|}\sum_{t\in W_m}\ell_t^{\,i}$ for $m=1,\dots,M$.
  \ENDFOR
  \STATE For each $m$, set $\ell_m^{\min}\leftarrow\min_{j\in\mathcal{A}}\tilde{\ell}_m^{\,j}$ and $\ell_m^{\max}\leftarrow\max_{j\in\mathcal{A}}\tilde{\ell}_m^{\,j}$.

  \FOR{$i \in \mathcal{A}$}
    \STATE \textbf{Stage 2:} Set $\alpha_{t,i}\leftarrow\alpha_t^{\mathrm{base}}$ for all $t\in W_1$.
    \FOR{$m=2,\dots,M$}
      \STATE $t_m \leftarrow \min(mK_{\mathrm{win}},T)$,\quad $t_{m-1}\leftarrow(m-1)K_{\mathrm{win}}$.
      \STATE $\tilde{\alpha}_{m,i}\leftarrow \mathrm{clip}\!\big(\Psi_m(\tilde{\ell}_m^{\,i}),\,\alpha_{\min},\,1\big)$ via Eq.~\ref{eq:window_mapping} (uses $\alpha_{t_m}^{\mathrm{base}},\alpha_{t_{m-1}}^{\mathrm{base}}$).
      \STATE Set $\alpha_{t,i}\leftarrow\tilde{\alpha}_{m,i}$ for all $t\in W_m$.
    \ENDFOR
    \STATE Reconstruct $\bar{\alpha}^{\,i}_{0}\leftarrow 1$;
    $\bar{\alpha}^{\,i}_{t}\leftarrow\bar{\alpha}^{\,i}_{t-1}\cdot\alpha_{t,i}$ for $t=1,\dots,T$.
  \ENDFOR
\ENDIF

\STATE \textbf{return} $\{\bar{\alpha}^{\,i}_{t}\}_{t=1}^T$.
\end{algorithmic}
\end{algorithm}

\subsection{Worked Toy Example: Graph-Aware Adaptive Noising}
\label{app:toy_example}

\noindent\textbf{Goal.}\label{app:toy-goal}
This toy example demonstrates Algorithm~\ref{alg:graph_aware_training} end-to-end on a small KG and its serialized
sequence. We illustrate (i) how token-wise denoising difficulty can exhibit minor non-monotone fluctuations over
diffusion time, (ii) how Algorithm~\ref{alg:graph_aware_training} maps \emph{window-averaged} difficulty to
\emph{per-step} noising coefficients via Eq.~\ref{eq:window_mapping}, and (iii) how the reconstructed token-wise
cumulative schedule preserves more signal for a hard, graph-aligned token at late (high-noise) steps.
We use full diffusion timesteps $T=2000$ and report values on a sampled grid
$\mathcal{T}=\{1,100,300,600,900,1200,1500,1700,1850,2000\}$ for readability; all window averages and schedule
construction are performed over all $t\in\{1,\dots,T\}$ as in the actual method.

\textbf{Toy KG and serialized sequence.}\label{app:toy-kg}
Consider a small KG in triplet form:
\begin{equation}
\label{eq:toy_kg_triplets}
\mathbf{G}=
\Big\{
(\texttt{Interstellar},\texttt{directed\_by},\texttt{Christopher\_Nolan}),
(\texttt{Interstellar},\texttt{release\_year},\texttt{2014})
\Big\}.
\end{equation}
A corresponding serialization is:
\begin{equation}
\label{eq:toy_sequence}
\mathbf{S}=
[\texttt{Interstellar},\ \texttt{directed},\ \texttt{by},\ \texttt{Christopher},\ \texttt{Nolan},\ \texttt{(2014)}].
\end{equation}
Assume graph--text alignment yields
$\mathcal{A}=\{\texttt{Interstellar},\texttt{Christopher},\texttt{Nolan}\}$.
We focus on a hard aligned token $i=\texttt{Christopher}$; unaligned tokens $i\notin\mathcal{A}$ retain the
baseline schedule as in Algorithm~\ref{alg:graph_aware_training}.

\paragraph{Baseline schedule (Algorithm~\ref{alg:graph_aware_training}: \texttt{REQUIRE}).}\label{app:toy-baseline}
We use a sqrt baseline cumulative schedule
$\bar{\alpha}_t = (1 - t/T)^2$ with $\bar{\alpha}_0=1$.
Table~\ref{tab:toy_baseline} reports sampled baseline cumulative values and baseline SNR
$\mathrm{SNR}^{\mathrm{base}}_t=\bar{\alpha}_t/(1-\bar{\alpha}_t)$.
We also use baseline per-step coefficients $\alpha_t^{\mathrm{base}}$; when reporting window endpoints
$(t_{m-1},t_m)$ we use the geometric mean
$\alpha^{\mathrm{base}}_{t_m}=(\bar{\alpha}_{t_m}/\bar{\alpha}_{t_{m-1}})^{1/(t_m-t_{m-1})}$,
which is equivalent to the per-step coefficients within that window.

\begin{table}[h!]
\caption{Baseline cumulative schedule and baseline SNR on $\mathcal{T}$.}
\label{tab:toy_baseline}
\centering
\small
\begin{tabular}{rcccccccccc}
\toprule
$t \in \mathcal{T}$ & 1 & 100 & 300 & 600 & 900 & 1200 & 1500 & 1700 & 1850 & 2000 \\
\midrule
$\bar{\alpha}_t$ &
0.999 & 0.980 & 0.940 & 0.880 & 0.780 & 0.650 & 0.520 & 0.430 & 0.360 & 0.300 \\
$\mathrm{SNR}^{\mathrm{base}}_t$ &
999.0 & 49.0 & 15.7 & 7.33 & 3.55 & 1.86 & 1.08 & 0.75 & 0.56 & 0.43 \\
\bottomrule
\end{tabular}
\end{table}

\paragraph{Stage 1: token-wise difficulty.}\label{app:toy-stage1}
Algorithm~\ref{alg:graph_aware_training} estimates the denoising difficulty
$\{\ell_t^{\,i}\}_{t=1}^{T}$ for each aligned token.
Denoising generally becomes hardest at late steps, while intermediate steps may exhibit minor fluctuations.
Table~\ref{tab:toy_loss} shows a sampled profile for the hard token.

\begin{table}[h!]
\caption{Sampled Stage-1 difficulty profile for $i=\texttt{Christopher}$.}
\label{tab:toy_loss}
\centering
\small
\begin{tabular}{rcccccccccc}
\toprule
$t \in \mathcal{T}$ & 1 & 100 & 300 & 600 & 900 & 1200 & 1500 & 1700 & 1850 & 2000 \\
\midrule
$\ell_t^{\,i}$ &
0.020 & 0.030 & 0.050 & 0.082 & 0.078 & 0.115 & 0.160 & 0.155 & 0.185 & 0.200 \\
\bottomrule
\end{tabular}
\end{table}

\paragraph{Stage 2: window-averaged mapping to per-step coefficients.}\label{app:toy-stage2}
Algorithm~\ref{alg:graph_aware_training} performs loss-to-noise mapping over non-overlapping diffusion windows
$W_m$ of length $K_{\mathrm{win}}=200$.
For each window $W_m$, we compute the window-averaged difficulty
$\tilde{\ell}_m^{\,i}=\frac{1}{|W_m|}\sum_{t\in W_m}\ell_t^{\,i}$.
We apply Eq.~\ref{eq:window_mapping} to obtain
$\tilde{\alpha}_{m,i}=\mathrm{clip}(\Psi_m(\tilde{\ell}_m^{\,i}),\alpha_{\min},1)$,
and set $\alpha_{t,i}=\tilde{\alpha}_{m,i}$ for all $t\in W_m$.
For the first window $W_1$, no loss-based adaptation is applied and
$\alpha_{t,i}=\alpha_t^{\mathrm{base}}$.

For this toy, we assume window-wise extrema over aligned tokens
$\ell_m^{\min}=0.03$ and $\ell_m^{\max}=0.20$, consistent with the sampled Stage-1 range.

\begin{table}[h!]
\caption{Window-level mapping for $i=\texttt{Christopher}$ using a sqrt baseline schedule.}
\label{tab:toy_windows}
\centering
\small
\begin{tabular}{rccccc}
\toprule
Window $m$ & $(t_{m-1},t_m)$ &
$\alpha^{\mathrm{base}}_{t_{m-1}}$ &
$\alpha^{\mathrm{base}}_{t_m}$ &
$\tilde{\ell}_m^{\,i}$ &
$\tilde{\alpha}_{m,i}$ \\
\midrule
1  & (0,200)     & --     & --     & --     & $\alpha_t^{\mathrm{base}}$ \\
2  & (200,400)   & 0.9989 & 0.9986 & 0.045  & 0.9988 \\
4  & (600,800)   & 0.9984 & 0.9980 & 0.080  & 0.9983 \\
5  & (800,1000)  & 0.9980 & 0.9976 & 0.095  & 0.9979 \\
7  & (1200,1400) & 0.9971 & 0.9962 & 0.130  & 0.9969 \\
8  & (1400,1600) & 0.9962 & 0.9947 & 0.155  & 0.9958 \\
9  & (1600,1800) & 0.9947 & 0.9924 & 0.175  & 0.9941 \\
10 & (1800,2000) & 0.9924 & 0.9896 & 0.195  & 0.9918 \\
\bottomrule
\end{tabular}
\end{table}

\paragraph{Reconstructing the cumulative schedule.}\label{app:toy-reconstruct}
The token-wise cumulative schedule is reconstructed as
$\bar{\alpha}^{\,i}_{t,\mathrm{new}}=\prod_{k=1}^{t}\alpha_{k,i}$.
Table~\ref{tab:toy_final} reports sampled values.

\begin{table}[h!]
\caption{Baseline vs.\ reconstructed token-wise cumulative schedule for $i=\texttt{Christopher}$.}
\label{tab:toy_final}
\centering
\small
\begin{tabular}{rcccccccccc}
\toprule
$t \in \mathcal{T}$ & 1 & 100 & 300 & 600 & 900 & 1200 & 1500 & 1700 & 1850 & 2000 \\
\midrule
Baseline $\bar{\alpha}_t$ &
0.999 & 0.980 & 0.940 & 0.880 & 0.780 & 0.650 & 0.520 & 0.430 & 0.360 & 0.300 \\
Adaptive $\bar{\alpha}^{\,i}_{t,\mathrm{new}}$ &
0.999 & 0.982 & 0.947 & 0.892 & 0.804 & 0.683 & 0.555 & 0.462 & 0.392 & 0.300 \\
\bottomrule
\end{tabular}
\end{table}

\paragraph{SNR comparison.}\label{app:toy-snr}
Define token-wise SNR as
$\mathrm{SNR}_{t,i}=\bar{\alpha}^{\,i}_{t,\mathrm{new}}/(1-\bar{\alpha}^{\,i}_{t,\mathrm{new}})$.
Table~\ref{tab:toy_snr} compares baseline and adaptive SNR.

\begin{table}[h!]
\caption{Baseline vs.\ adaptive SNR for the aligned token (Factual tokens).}
\label{tab:toy_snr}
\centering
\small
\begin{tabular}{rccccccc}
\toprule
$t$ & 600 & 900 & 1200 & 1500 & 1700 & 1850 & 2000 \\
\midrule
Baseline SNR &
7.33 & 3.55 & 1.86 & 1.08 & 0.75 & 0.56 & 0.43 \\
Adaptive SNR &
8.25 & 4.10 & 2.15 & 1.25 & 0.86 & 0.64 & 0.43 \\
\bottomrule
\end{tabular}
\end{table}

\subsection{Training Details and Hyper-parameters}\label{App:Train}
\textit{Model variants}:
We train two Transformer–based denoisers:  
(i) a 6-encoder / 6-decoder architecture with \(\approx 50\) M parameters, and  
(ii) a 6-encoder / 9-decoder architecture with \(\approx 63\) M parameters.  
Both use GeLU activations and share all other hyper-parameters.

\textit{Diffusion setup}:
A fixed diffusion horizon of \(T=2000\) timesteps is employed, following the \textit{sqrt} noise schedule introduced in DiffusionLM~\cite{li2022diffusionlmimprovescontrollabletext}.  
Inputs are tokenised with the \texttt{bert-base-uncased}  vocabulary~\cite{devlin-etal-2019-bert}.  
The graph-aware noising schedule is calculated every 20,000 training steps.

\textit{Optimization}:
All experiments use AdamW with a peak learning rate of \(1\!\times\!10^{-4}\), a linear warm-up of 10,000 steps, and linear decay to zero.  
Gradient norms are clipped to \(1.0\); no label-smoothing or dropout is applied beyond the architectural dropout already reported in the main text.

\textit{Training setup}:
Each model is trained for up to 200,000 steps per dataset:\\
(1) The 50 M model achieves its best validation metrics after \(\sim\!190,000\) steps.\\
(2) The 63 M model converges at the full 200,000-steps budget.

These numbers were found to be stable across all datasets considered.

\noindent\textbf{Reference.}
All reproducibility-critical settings are summarized in
Table~\ref{tab:hyperparams_train} (training + optimization + alignment + graph-aware schedule; cf.\ Algorithm~1 and Eq.~\ref{eq:window_mapping})
and Table~\ref{tab:hyperparams_infer} (inference/sampling protocol, metric settings for FGT/ESR, and mapping-function ablations).
We refer readers to Table~\ref{tab:hyperparams_train} for the exact values of $(T, K_{\text{up}}, K_{\text{win}}, \tau)$ and to
Table~\ref{tab:hyperparams_infer} for $(\lambda, T')$ and the DDIM sampling framework.

\begin{table}[h!]
\centering
\small
\renewcommand{\arraystretch}{1.25}
\setlength{\tabcolsep}{6pt}
\caption{
\textbf{Training, optimization, alignment, and graph-aware schedule hyperparameters for \texttt{DLM4G}.}
This table reports all hyperparameters required to reproduce training of the encoder--decoder denoiser
and the graph-aware adaptive noising schedule (Stage~1/2, Algorithm~1).
We distinguish between (i) the fixed token-level sequence length used for diffusion and padding,
and (ii) the variable word-level sequence length \(|S|\) used only in evaluation metrics (FGT/ESR).
Graph-aware schedules are learned only for aligned token positions \(i \in \mathcal{A}\),
where \(\mathcal{A}\) is obtained from an offline entity/relation alignment pipeline (Appendix~\ref{app:alignment-analysis}).
}
\label{tab:hyperparams_train}
\rowcolors{2}{BandBlue}{white}
\begin{tabular}{llp{6.55cm}}
\toprule
\rowcolor{HeaderBlue}
\textbf{Category} & \textbf{Hyperparameters} & \textbf{Value} \\
\midrule

Model & Denoiser variants &
(i) 6-encoder / 6-decoder (\(\approx\)50M params);
(ii) 6-encoder / 9-decoder (\(\approx\)63M params) \\
Model & Activation & GeLU \\
\midrule

Diffusion &
Diffusion timesteps \((T)\) & \(2000\) \\
Diffusion &
Baseline cumulative schedule \(\big(\bar{\alpha}^{\mathrm{base}}_t\big)\) &
Sqrt schedule (DiffusionLM-style) \\
Diffusion &
Token-wise schedules \(\big(\bar{\alpha}^{\,i}_{t,\mathrm{new}}\big)\) &
Learned only for aligned positions \(i \in \mathcal{A}\);
unaligned \(i \notin \mathcal{A}\) retain \(\bar{\alpha}^{\mathrm{base}}_t\) \\
\midrule

Graph-aware schedule &
Update interval \((K_{\text{up}})\) &
Every \(20{,}000\) training steps \\
Graph-aware schedule &
Window length \((K_{\text{win}})\) &
\(200\) diffusion steps (non-overlapping windows) \\
Graph-aware schedule &
Loss-to-noise mapping \(\big(\Psi_i(\cdot)\big)\) &
Linear mapping (Eq.\ref{eq:window_mapping}); polynomial / cosine used only in ablations \\
Graph-aware schedule &
Stabilizer \((\tau)\) &
\(10^{-8}\) (numerical stability in Eq.(\ref{eq:window_mapping})) \\
\midrule

Alignment &
Alignment set \((\mathcal{A})\) &
Training-only set of non-\texttt{[PAD]} token positions
aligned to KG entities/relations (Appendix~\ref{app:alignment-analysis}) \\
Alignment &
Alignment granularity &
Token-level (BERT wordpieces); evaluation entities extracted at word level \\
Alignment &
Max aligned span length &
Up to contiguous entity/relation mentions; no span merging across \texttt{[SEP]} \\
\midrule

Tokenization &
Tokenizer / vocab &
\texttt{bert-base-uncased} with added \texttt{[HEAD]}, \texttt{[REL]}, \texttt{[TAIL]}, \texttt{[SEP]} \\
\midrule

Sequence length &
Max output length (tokens) &
\(64\) tokens (padding/truncation for training) \\
Sequence length &
Word-level length \((|S|)\) &
\(N = 16\) words (used only in FGT/ESR) \\
\midrule

Optimization &
Optimizer &
AdamW \\
Optimization &
Peak learning rate &
\(1\times 10^{-4}\) \\
Optimization &
Warmup &
\(10{,}000\) steps (linear) \\
Optimization &
LR decay &
Linear decay to zero \\
Optimization &
Gradient clipping &
Global norm clip \(=1.0\) \\
Regularization &
Dropout rate &
\(0.2\) (Transformer layers) \\
Regularization &
Label smoothing &
None \\
\midrule

Training length &
Training steps &
\(200{,}000\) steps per dataset \\
Training setup &
Batch size &
\(128\) \\
\bottomrule
\end{tabular}
\end{table}

\begin{table}[t]
\centering
\small
\renewcommand{\arraystretch}{1.25}
\setlength{\tabcolsep}{6pt}
\caption{
\textbf{Inference, metric, and ablation hyperparameters.}
We report the settings used for (i) the proposed evaluation metrics (FGT and ESR),
(ii) inference-time sampling and efficiency studies, and (iii) mapping-function ablations.
FGT is reported at the default hallucination-penalty tradeoff \(\lambda=0.5\);
ESR has no tunable hyperparameters.
Efficiency experiments vary the number of DDIM denoising steps \(T'\) and use a fixed timing protocol.
}
\label{tab:hyperparams_infer}
\rowcolors{2}{BandBlue}{white}
\begin{tabular}{llp{6.55cm}}
\toprule
\rowcolor{HeaderBlue}
\textbf{Category} & \textbf{Hyperparameter (symbol)} & \textbf{Value} \\
\midrule

Metrics & FGT tradeoff \((\lambda)\) & Default: \(\lambda=0.5\) \\
Metrics & ESR hyperparameters & None \\
\midrule

Inference (schedule) & Attention weight \(\big(w_i^t\big)\) & Total normalized decoder-to-graph cross-attention mass (averaged across heads) \\
Inference (schedule) & Anchor schedule \(\big(\bar{\alpha}^{\mathrm{anchor}}_t\big)\) & Average of learned \(\bar{\alpha}^{\,i}_{t,\mathrm{new}}\) over aligned (non-\texttt{[PAD]}) training occurrences \\
Inference (schedule) & Blended schedule \(\big(\bar{\alpha}^{\,i}_t\big)\) & \(\bar{\alpha}^{\,i}_t=(1-w_i^t)\bar{\alpha}^{\mathrm{base}}_t+w_i^t\bar{\alpha}^{\mathrm{anchor}}_t\) \\
\midrule

Inference (efficiency) & Decoding & Greedy (beam size 1); same max length + early stopping across methods \\
Inference (efficiency) & Sampler & DDIM sampling \\
Inference (efficiency) & DDIM steps \((T')\) & \(T' \in \{2000,1000,500,100\}\) \\
Inference (efficiency) & Timing protocol & Batch size 50; single NVIDIA V100; report total wall-clock time \\
\midrule

Mapping ablations & Mapping family \(\big(\Psi_i(\cdot)\big)\) & Linear / polynomial / cosine (schedule assignment ablation) \\
Mapping ablations & Shape function \(\big(\phi(\cdot)\big)\) & \(\phi_{\text{lin}}(u)=u\); \(\phi_{\text{poly}}(u)=u^p\) with \(p=2\); \(\phi_{\exp}(u)=\frac{e^{\beta u}-1}{e^\beta-1}\) with \(\beta=3\); \(\phi_{\cos}(u)=\tfrac{1}{2}(1-\cos(\pi u))\) \\
\bottomrule
\end{tabular}
\end{table}

\subsection{SoTA G2S Baselines comparison}\label{app:G2S_sota}

\noindent\textbf{Results Overview.}
Table~\ref{full table} reports a side-by-side comparison with prior SoTA graph-to-sequence (G2S) baselines on
\textbf{GenWiki} and \textbf{TekGEN}. We include both \texttt{DLM4G}-1.o and \texttt{DLM4G}-2.o variants, and highlight
the best baseline (underlined) and our best result (bold). The last row (\%Gain) summarizes relative improvement of
\texttt{DLM4G}-2.o over the strongest baseline for each metric.

\begin{table}[H]
\caption{ Performance of \texttt{DLM4G} compared with baselines on (a) GenWiki and (b) TekGEN.}
  \label{full table}
  \scriptsize 
  \centering
  \captionsetup[subtable]{justification=centering}
  \begin{subtable}[t]{0.5\linewidth}
    \centering
    \setlength{\tabcolsep}{3.5pt}
    \begin{tabular}{clccccc}
      \toprule
      & & \multicolumn{5}{c}{\textbf{GenWiki}} \\
      \cmidrule(lr){3-7}
      & \textbf{Baselines} & \textbf{B} & \textbf{CrF++} & \textbf{M} & \textbf{B-F1} & \textbf{MVE} \\
      \midrule
      & Rule-Based      & 0.219 & 0.360 & 0.397 & 0.679 & 0.822 \\
      & Direct-Transfer   & 0.234 & 0.483 & 0.332 & 0.808 & 0.801 \\
      & Noisy-Sup.      & 0.384 & 0.623 & 0.414 & \underline{0.878} & \textbf{0.901} \\
      \rowcolor{white!70!yellow}
      & \texttt{DLM4G}-1.o & \underline{0.401} & \underline{0.663} & \underline{0.527} & 0.857 & 0.841 \\
      \rowcolor{white!70!yellow}
      & \texttt{DLM4G}-2.o & \textbf{0.469} & \textbf{0.748} & \textbf{0.574} & \textbf{0.899} & \underline{0.892} \\
      \rowcolor{white!70!gray}
      & \%Gain          & \textbf{+22.1\%}& \textbf{+20.0\%} & \textbf{+38.6\%} & \textbf{+2.4\%} & 0.0\% \\
      \bottomrule
    \end{tabular}
  \end{subtable}\hfill
  \begin{subtable}[t]{0.49\linewidth}
    \centering
    \setlength{\tabcolsep}{3pt}
    \begin{tabular}{clccccc}
      \toprule
      & & \multicolumn{5}{c}{\textbf{TekGEN}} \\
      \cmidrule(lr){3-7}
      & \textbf{Baselines} & \textbf{B} & \textbf{CrF++} & \textbf{M} & \textbf{B-F1} & \textbf{MVE} \\
      \midrule
      & Rule-based        & 0.189 & 0.309 & 0.301 & 0.509 & 0.672 \\
      & ReGen-SCST      & 0.219 & 0.385 & 0.223 & 0.698 & 0.719 \\
      & ReGen-CE        & 0.199 & 0.372 & 0.214 & 0.612 & 0.701 \\
      \rowcolor{white!70!yellow}
      & \texttt{DLM4G}-1.o & \underline{0.247} & \underline{0.493} & \underline{0.375} & 0.795 & 0.781 \\
      \rowcolor{white!70!yellow}
      & \texttt{DLM4G}-2.o & \textbf{0.253}  & \textbf{0.522} & \textbf{0.414} & \textbf{0.847} & \textbf{0.820} \\
      \rowcolor{white!70!gray}
      & \%Gain          & \textbf{+10.9\%} & \textbf{+41.1\%} & \textbf{+96.2\%} & \textbf{+21.3\%} & \textbf{+14.0\%} \\
      \bottomrule
    \end{tabular}
  \end{subtable}
\end{table}

\subsection{Mapping Function Ablations}\label{schedules}

\begin{figure*}[t!]
    \centering
    \includegraphics[width=\linewidth]{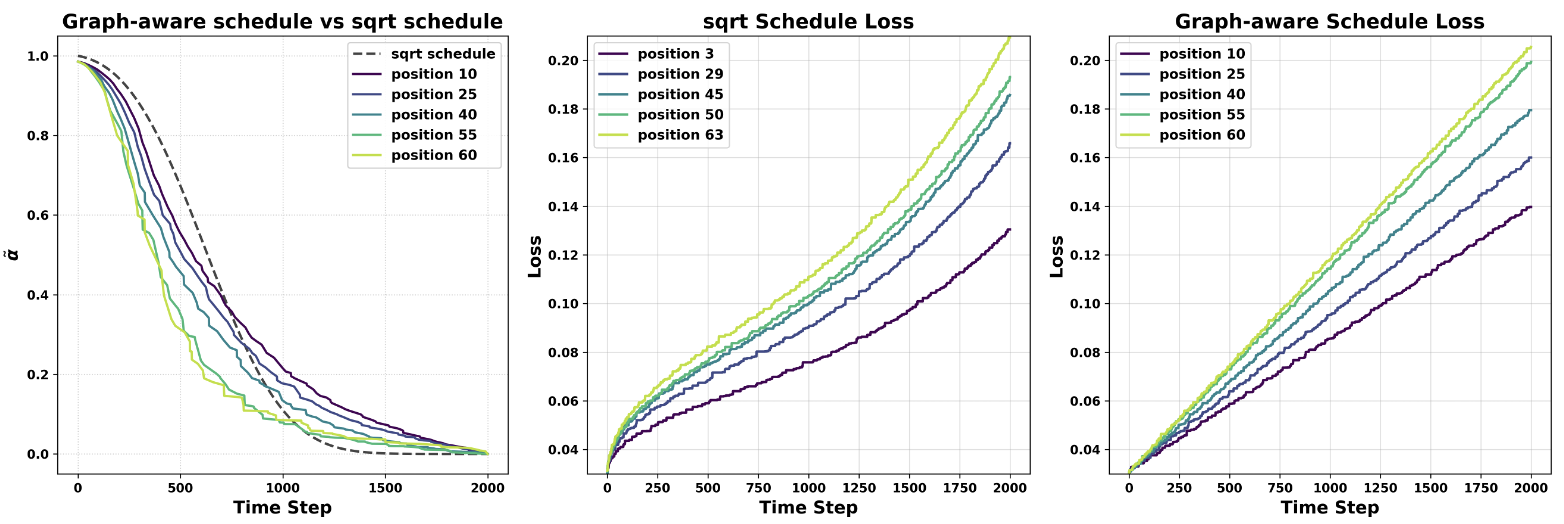}
    \caption{(\textbf{L}eft) Noise schedule corresponding to alignment set $i \in \mathcal{A}$ (position 10, 25,.., 60) compared against \textit{sqrt} schedule for $i \notin \mathcal{A}$ (position 3, 29,.., 63); (\textbf{M}id) The loss profile across time steps ($t=0 \rightarrow T$) for the syntactic tokens (\textbf{R}ight) The loss profile across time steps ($t=0 \rightarrow T$) for the factual tokens.}
    
    \label{sqrt}
\end{figure*}

\textbf{Choice of Mapping Function:} As discussed in Sec.~\ref{sec:graph_aware_noising}, the graph-aware schedule
is obtained by mapping token-wise difficulty profiles
$\{\ell_t^{\,i}\}_{t=1}^T$ through a monotone mapping function to obtain token-wise coefficients.
We define the mapping \emph{piecewise} over intervals indexed by $m$ with endpoints $(t_{m-1}, t_m)$.
Let
\[
\ell_m^{\min} = \min\!\big(\ell_{t_{m-1}}^{\,i}, \ell_{t_m}^{\,i}\big),
\qquad
\ell_m^{\max} = \max\!\big(\ell_{t_{m-1}}^{\,i}, \ell_{t_m}^{\,i}\big),
\]
and for $x \in [\ell_m^{\min}, \ell_m^{\max}]$ define
\begin{equation}
\label{eq:mapping_piecewise}
\Psi_m(x)
=
\alpha_{t_m}^{\mathrm{base}}
+
\phi\!\left(
\frac{x-\ell_m^{\min}}{\ell_m^{\max}-\ell_m^{\min}+\tau}
\right)
\Big(\alpha_{t_{m-1}}^{\mathrm{base}}-\alpha_{t_m}^{\mathrm{base}}\Big),
\qquad m=2,\dots,M,
\end{equation}
where $\tau>0$ is a stabilizer (avoids division by very small $\ell_m^{\max}-\ell_m^{\min}$).
In the main experiments we use a linear mapping $\phi_{\text{lin}}(u)=u$, which recovers Eq.~(3).
Here we ablate three smooth alternatives: exponential, cosine, and polynomial mappings, all with the same monotonicity and endpoint behavior.

\begin{figure}[t!]
\centering
\begin{minipage}[h!]{0.31\textwidth}
  \centering
  \includegraphics[width=\linewidth]{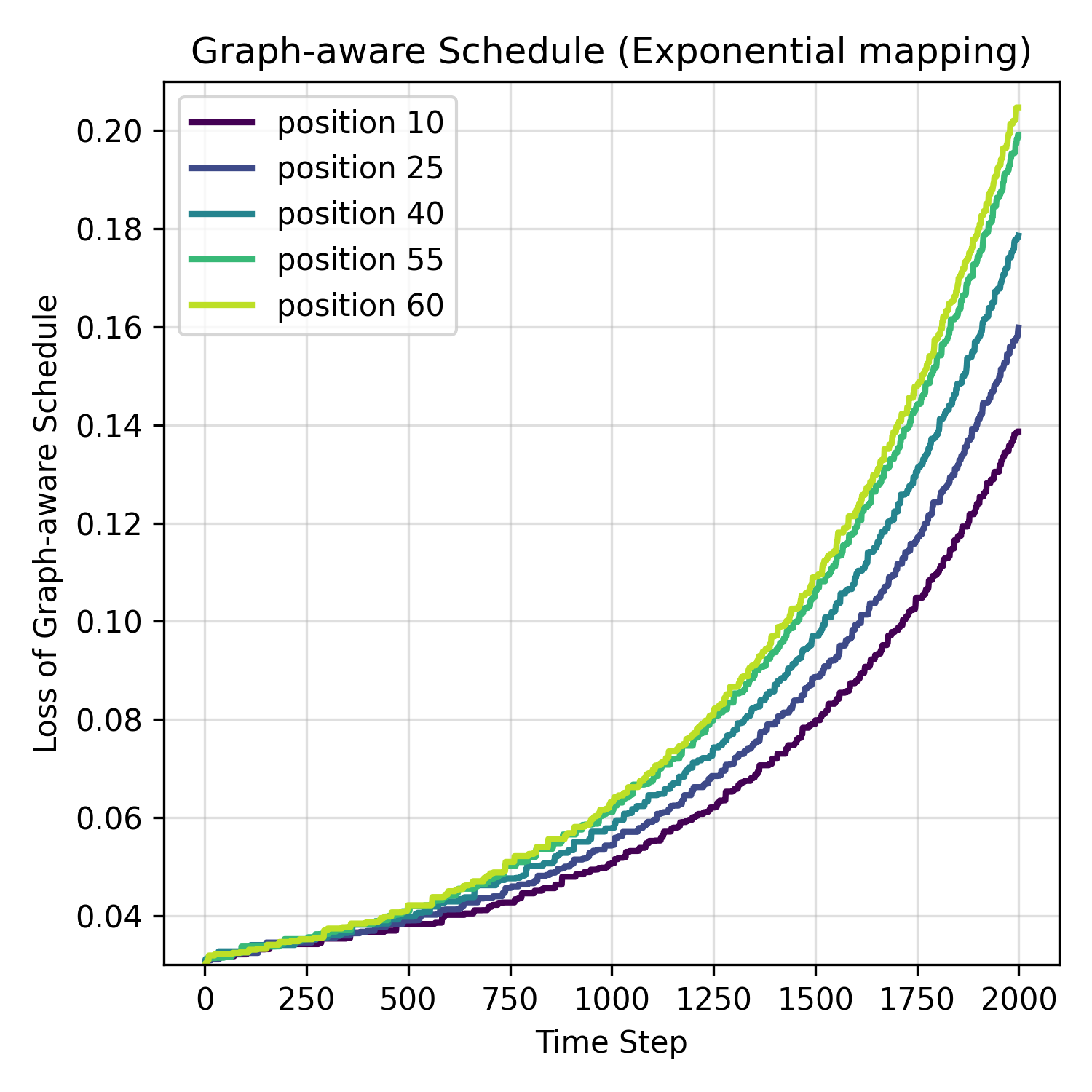}
\end{minipage}\hfill
\begin{minipage}[h!]{0.31\textwidth}
  \centering
  \includegraphics[width=\linewidth]{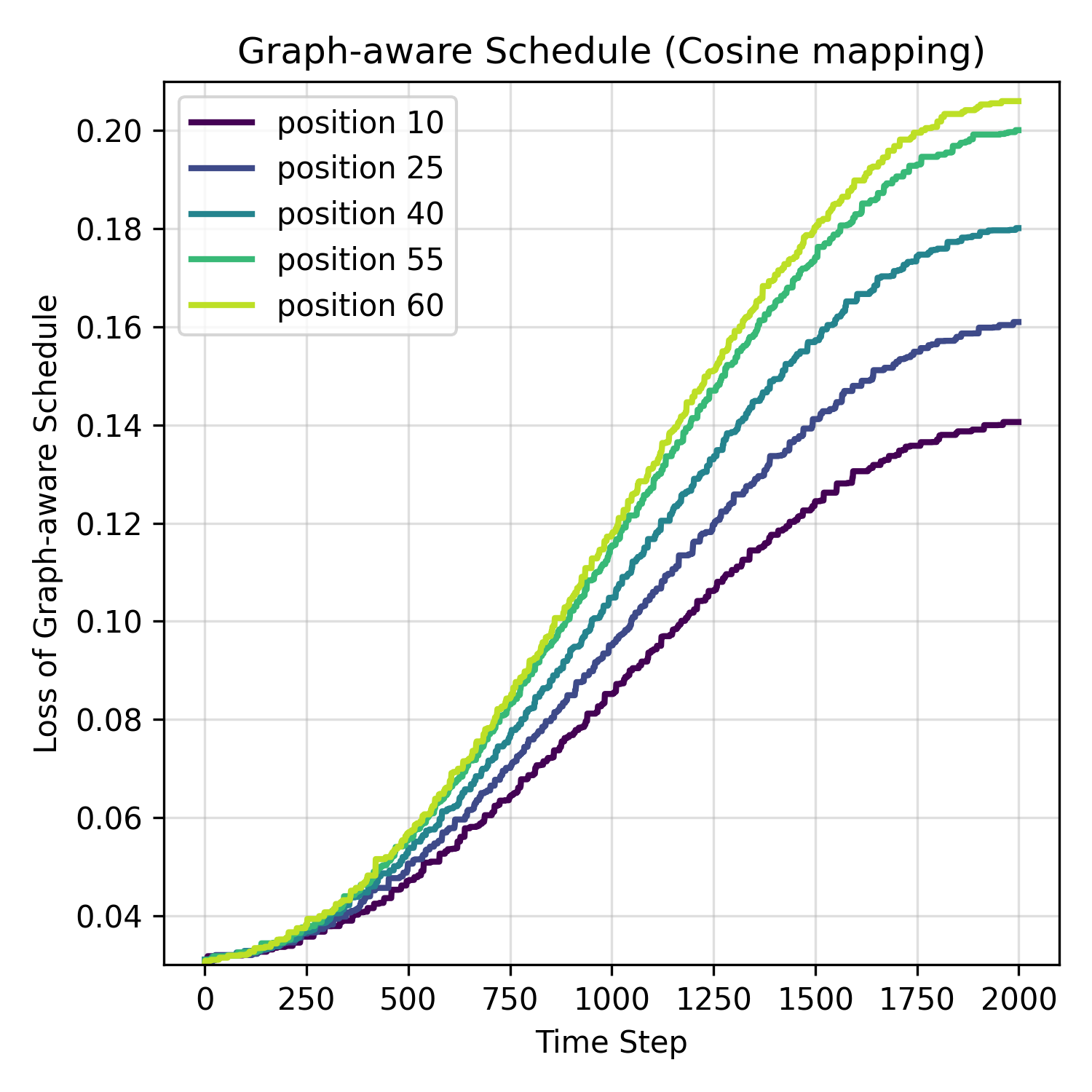}
\end{minipage}\hfill
\begin{minipage}[h!]{0.31\textwidth}
  \centering
  \includegraphics[width=\linewidth]{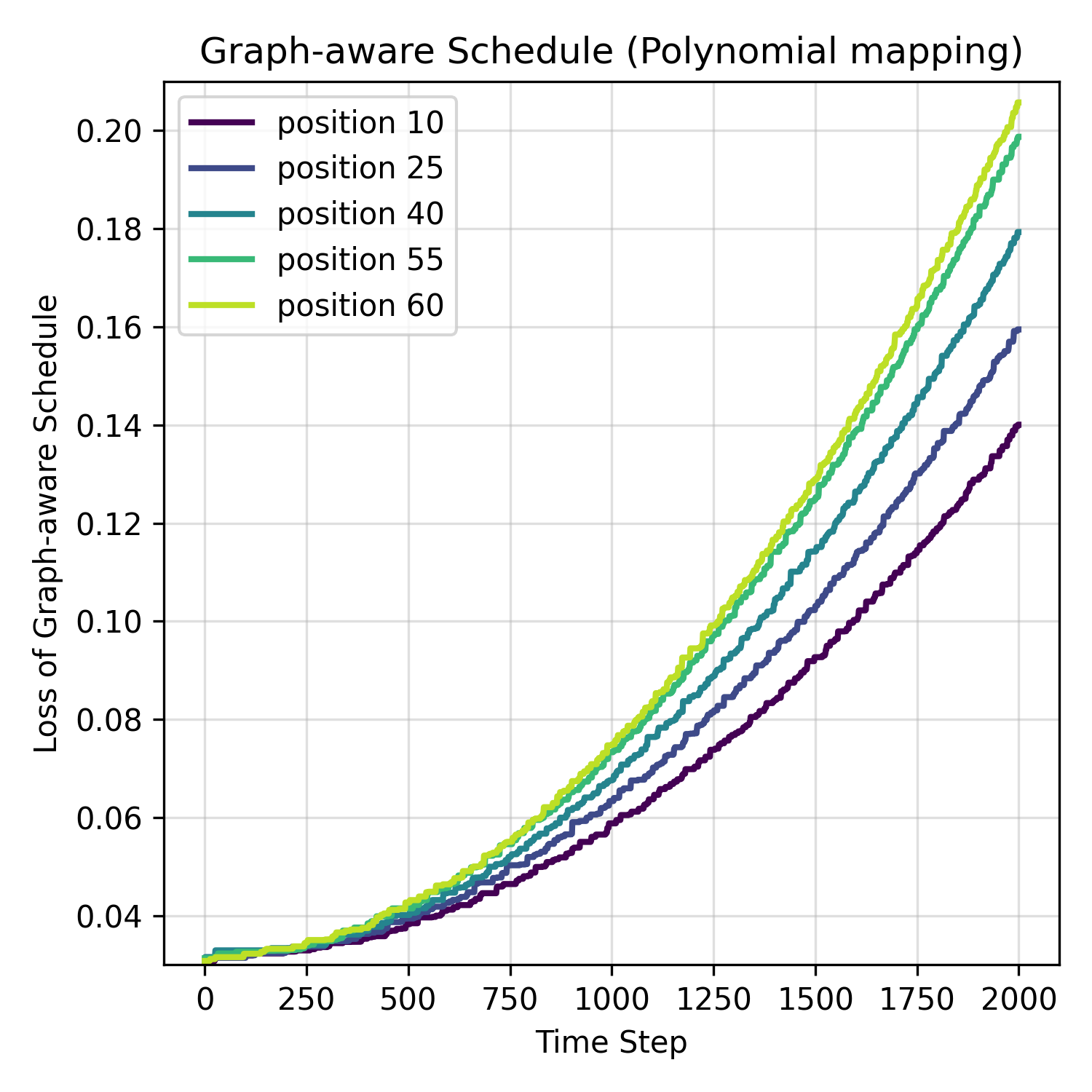}
\end{minipage}\hfill
\caption{Induced loss profiles for the graph-aware
schedule under different mappings: (\textbf{L}eft) exponential,
(\textbf{M}id) cosine, and (\textbf{R}ight) polynomial. Curves show loss trajectories for
aligned token positions at steps 10, 25, 40, 55, and 60.}
\label{fig:dlm4g_schedule_ablation}
\end{figure}

Table~\ref{tab:mapping_functions} summarizes the functional forms, and
Fig.~\ref{fig:dlm4g_schedule_ablation}(\textbf{L}--\textbf{R}) visualizes the induced loss
profiles over diffusion steps for aligned token positions
(10, 25, 40, 55, 60).
All three mappings introduce sharper non-linearities (e.g., flatter early
regions and steeper tails), which make the token-wise loss grow more
abruptly near the end of the diffusion process. Empirically, this
concentration of noise updates degrades downstream performance: we observe
small but consistent drops in \textbf{BLEU} (see Table~\ref{ablation}). Hence, we retain the simple linear mapping
$\phi_{\mathrm{lin}}(u)=u$ in \texttt{DLM4G}.

\textbf{Performance:} In addition to BLEU, we assess how the choice of mapping function affects
factual grounding and edit sensitivity on WikiOFGraph, using the metrics
introduced in Sec.~\ref{facts}.
For FGT we report FGT@$\lambda=0.5$, our default setting which balances the
penalty on hallucinated entities; ESR has no hyperparameter.
Table~\ref{tab:noise_mapping_ablation} extends the mapping ablation to these
metrics.
We find that the graph-aware linear mapping over aligned tokens
$\mathcal{A}$ improves BLEU and FGT while also achieving the highest ESR,
whereas the more non-linear polynomial and cosine mappings consistently
degrade all three metrics.
This supports our qualitative analysis in
Fig.~\ref{fig:dlm4g_schedule_ablation} and further motivates our choice of
the linear mapping in \texttt{DLM4G}.
\begin{table}[h!]
\centering
\caption{Ablation of \texttt{DLM4G} noise schedules and mapping function
$\Psi_i(x)$ on WikiOFGraph.
We report BLEU (\textbf{B}), Factual Grounding (FGT@$\lambda=0.5$,
higher is better), and Edit Sensitivity Rate (ESR, higher is better).}
\label{tab:noise_mapping_ablation}
\setlength{\tabcolsep}{6pt}
\renewcommand{\arraystretch}{1.1}
\begin{tabular}{@{}lcccccc@{}}
\toprule
\texttt{DLM4G} & \textbf{Tokens} & \textbf{$\Psi_i(x)$} &
\textbf{B}& \textbf{FGT@0.5}  & \textbf{ESR} \\
\midrule
\rowcolor{white!70!yellow}
Graph-aware & $\mathcal{A}$ & linear &
\textbf{0.65} & \textbf{0.83} & \textbf{0.68} \\ 
\rowcolor{white!70!yellow}
Graph-aware & $\mathcal{A}$ & poly   &
0.61  & 0.80 & 0.61 \\        
\rowcolor{white!70!yellow}
Graph-aware & $\mathcal{A}$ & cosine &
0.62  & 0.80 & 0.63 \\        
\bottomrule
\end{tabular}
\end{table}

\begin{table}[h!]
\centering
\caption{Ablation of $\phi(u)$ used in the graph-aware schedule.}
\label{tab:mapping_functions}
\begin{tabular}{ll}
\toprule
Ablation      & $\phi(u)$ \\
\midrule
Linear            & $\phi_{\mathrm{lin}}(u) = u$ \\
Polynomial        & $\phi_{\mathrm{poly}}(u) = u^{p}$ \quad (we use $p = 2$) \\
Exponential       & $\phi_{\mathrm{exp}}(u) = \dfrac{e^{\beta u} - 1}{e^{\beta} - 1}$ \quad (we use $\beta = 3$) \\
Cosine            & $\phi_{\mathrm{cos}}(u) = \tfrac{1}{2}\bigl(1 - \cos(\pi u)\bigr)$ \\
\bottomrule
\end{tabular}
\end{table}

\subsection{Inference Efficiency Details}
\label{app:efficiency}

\textbf{Goal.}
We analyze the quality--efficiency tradeoff of \texttt{DLM4G} for graph-to-sequence generation by varying the number of
DDIM sampling steps $T'$ at inference time.

\textbf{Setup and timing protocol.}
We report total wall-clock inference time (seconds) for a fixed batch size of 50 on a single NVIDIA V100 GPU.
All methods use the same test split and identical decoding settings: greedy decoding (beam size 1) with the same
maximum generation length and early-stopping rule across models.
Unless stated otherwise, reported times include the full forward pass required to generate sequences from the input
graph representation under each method.

\textbf{DDIM sampling.}
For \texttt{DLM4G}, we use DDIM sampling with $T'$ denoising steps.
We vary $T'\in\{2000,1000,500,100\}$ for \texttt{DLM4G}-2.o while keeping all other inference hyperparameters fixed.

\textbf{Results.}
Table~\ref{tab:efficiency_breakdown} reports BLEU and wall-clock time.
At $T'=2000$, \texttt{DLM4G}-2.o achieves higher BLEU than diffusion counterparts while being substantially faster than
DiffuSeq and matching the decoding latency of SeqDiffuSeq.
As $T'$ decreases, runtime improves approximately linearly with $T'$ but quality degrades, yielding a clear crossover point:
\texttt{DLM4G} requires $\approx$1000 steps to surpass strong AR baselines under this evaluation setting.

\begin{table}[h]
\centering
\small
\renewcommand{\arraystretch}{1.15}
\setlength{\tabcolsep}{6pt}
\caption{
\textbf{Efficiency Analysis on WikiOFGraph.}
We compare \texttt{DLM4G} against autoregressive (AR) and diffusion baselines.
Time is measured as total inference wall-clock time for a \textbf{batch size of 50} on a single
\textbf{NVIDIA V100 GPU}. For \texttt{DLM4G}, we vary the number of DDIM denoising steps \(T'\).
We observe a crossover regime: \texttt{DLM4G} requires \(\approx\)1000 DDIM steps to surpass strong AR baselines in quality.
}
\vspace{0.4em}
\label{tab:efficiency_breakdown}
\rowcolors{2}{BandBlue}{white}
\begin{tabular}{lccccc}
\toprule
\rowcolor{HeaderBlue}
\textbf{Model} & \textbf{Params} & \textbf{DDIM Steps} & \textbf{Time (s)} & \textbf{Speedup} & \textbf{BLEU $\uparrow$} \\
\midrule

\rowcolor{GroupGray}
\multicolumn{6}{l}{\textit{\# Autoregressive Baselines}} \\
GPT-2 (Base) & 355M & 64 & $<$5s & -- & 0.285 \\
T5 (Small) & 60M & 64 & $<$5s & -- & 0.385 \\

\midrule
\rowcolor{GroupGray}
\multicolumn{6}{l}{\textit{\# Diffusion Baselines}} \\
DiffuSeq & 91M & 2000 & 317s & $1.0\times$ (Ref) & 0.628 \\
SeqDiffuSeq & 50M & 2000 & 89s & $3.6\times$ & 0.616 \\

\midrule
\rowcolor{GroupGray}
\multicolumn{6}{l}{\textit{\texttt{\# DLM4G} (Ours)}} \\
\rowcolor{OursGreen}
\texttt{DLM4G}-2.o & 63M & 2000 & 89s & $3.6\times$ & \textbf{0.654} \\
\rowcolor{OursGreen}
\texttt{DLM4G}-2.o & 63M & 1000 & 45s & $7.0\times$ & 0.551 \\
\rowcolor{OursGreen}
\texttt{DLM4G}-2.o & 63M & 500 & 23s & $13.8\times$ & 0.365 \\
\rowcolor{OursGreen}
\texttt{DLM4G}-2.o & 63M & 100 & 5s & $63.4\times$ & 0.312 \\

\bottomrule
\end{tabular}
\end{table}

\subsection{Zero-shot prompting}\label{App:FSZS}
\begin{figure}[ht!]
    \centering
    \includegraphics[width=\linewidth]{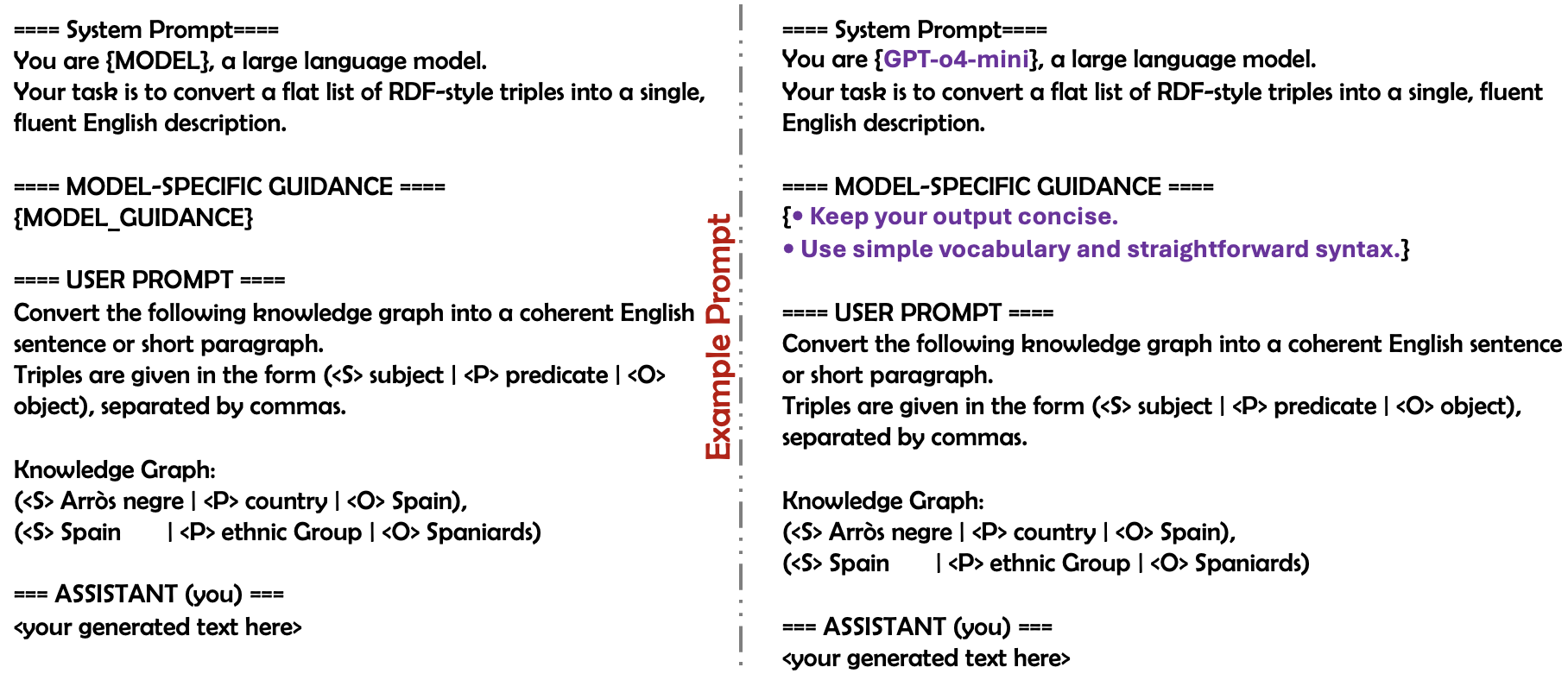}
    \caption{Zero-Shot Prompt Template for Knowledge Graph Verbalization Across Multiple LLMs}

    \label{fig:zero-shot}
\end{figure}

Zero-shot prompting (illustrated in Figure~\ref{fig:zero-shot}) exploits the rich, general‐purpose knowledge encoded in pretrained large language models (LLMs) to tackle novel tasks without additional fine-tuning. By casting tasks as natural-language instructions or templated prompts, models such as GPT-3 \cite{brown2020language}, DeepSeek \cite{deepseekai2024deepseekllmscalingopensource}, LLaMa-3 \cite{grattafiori2024llama3herdmodels}, and Qwen2.5 \cite{qwen2025qwen25technicalreport} demonstrate strong out-of-the-box performance across diverse applications. Prior work has shown that LLMs internalize extensive linguistic, factual, and procedural knowledge during self-supervised training, yielding robust zero-shot capabilities in text classification, machine translation \cite{raffel2020exploring}, and code generation \cite{chen2021evaluating}.
A typical zero-shot prompt comprises three components:  \\
(1) A \emph{system prompt} that assigns the model’s role (e.g., “You are \{MODEL\}, a large language model. Convert RDF triples into fluent English.”). \\
(2) A \emph{model-specific guidance} segment to steer style or brevity (e.g., “Keep your output concise.”).  \\
(3) A \emph{user prompt} presenting the task instance. 

For example: \textit{Convert the following knowledge graph into a single English sentence:}\\
$\langle S\rangle\ \text{Arròs negre}\ \langle P\rangle\ \text{country}\ \langle O\rangle\ \text{Spain},\;\langle S\rangle\ \text{Spain}\ \langle P\rangle\ \text{ethnic Group}\ \langle O\rangle\ \text{Spaniards}.$

In this study, we evaluate four models: DeepSeek (7 B), GPT-o4-mini (8 B), LLaMa-3 (8 B), and Qwen2.5 (7 B), to investigate how model scale, pretraining corpus, and architectural choices affect zero-shot generalization on knowledge-to-text tasks.

\subsection{Molecule Captioning}\label{App:mol}

\begin{figure}[t!]
  \centering
  \begin{subfigure}[b]{0.48\textwidth}
    \centering
    \includegraphics[width=0.8\linewidth]{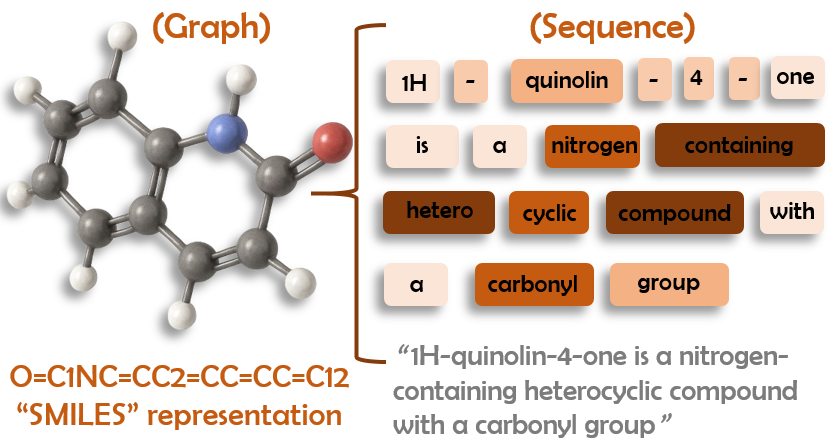}
    \label{fig:molcap-seq}
  \end{subfigure}
  \hfill
  \begin{subfigure}[b]{0.48\textwidth}
    \centering
    \includegraphics[width=0.7\linewidth]{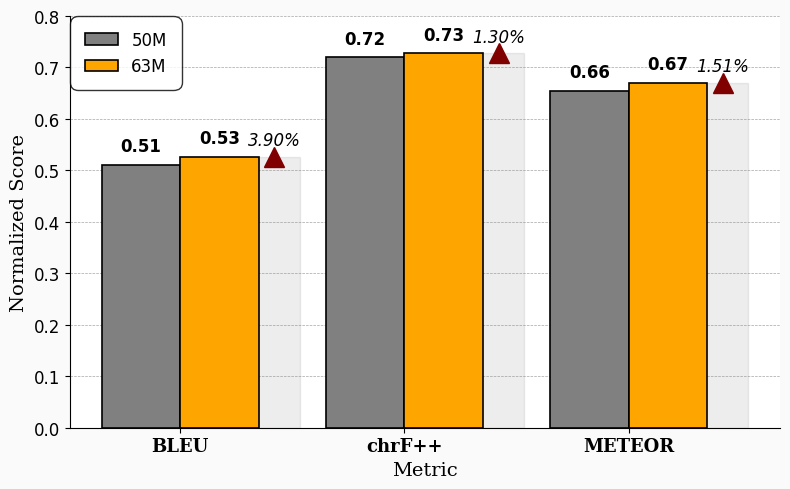}
    \label{fig:molcap-struct}
  \end{subfigure}
  \caption{Comparison of (left) framing molecule captioning as a G2S task and (right) the performance of \texttt{DLM4G}-1.o and \texttt{DLM4G}-2.o models on the molecule captioning dataset.}
  \label{molcap-overall}
\end{figure}

\begin{figure}[h!]
    \centering
    \includegraphics[width=1\linewidth]{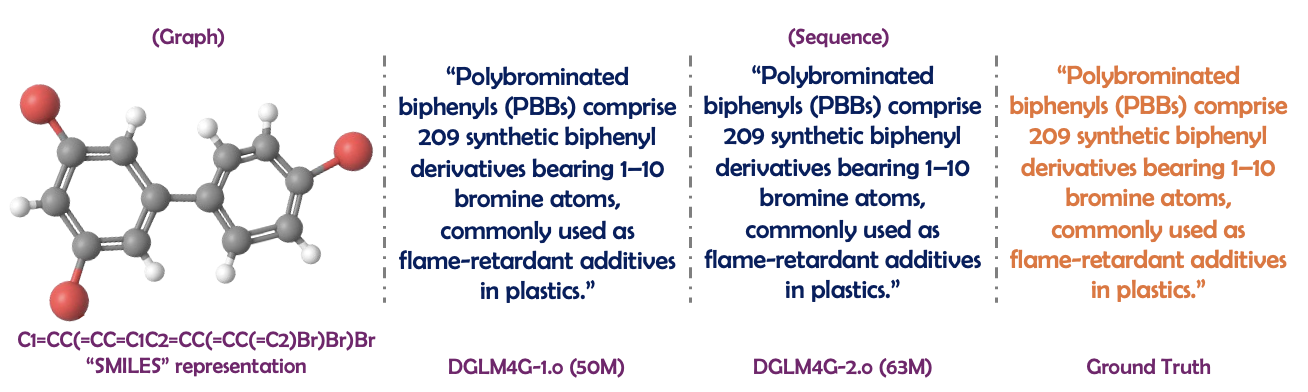}
    \caption{Qualitative assessment of molecule captioning by \texttt{DLM4G} given SMILES representations.}
    \label{finale}
\end{figure}

\noindent\textbf{Evaluation Summary:}
We include molecule captioning as an additional application of \texttt{DLM4G} beyond knowledge-grounded G2S.
Table~\ref{tab:single_dataset_comparison} reproduces the main quantitative comparison against strong baselines,
Fig.~\ref{molcap-overall} illustrates (left) the reduction of molecule captioning to a graph-/SMILES-to-text generation setting
and (right) the corresponding performance of our two model variants, and Fig.~\ref{finale} provides a qualitative example
with model outputs and the ground-truth caption for a representative molecule.
For full experimental details (data preprocessing, training protocol, and additional qualitative samples), see the main paper.

\begin{table}[h!]
    \centering
    \caption{Comparison of our \texttt{DLM4G} models against baselines.}
    \label{tab:single_dataset_comparison}
    \small
    \resizebox{0.7\linewidth}{!}{%
    \begin{tabular}{lccccccc}
    \toprule
    \textbf{Method} & \textbf{\#P} & \textbf{B} & \textbf{CrF++} & \textbf{M} & \textbf{B-F1} & \textbf{MVE}  \\
    \midrule
    MolT5 (B) & 220M & 0.452 & 0.651 & 0.510 & 0.681 & 0.852 \\
    GitMol & 700M & 0.475 & 0.680 & 0.532 & 0.751 & 0.875 \\
    GraphT5 & 272M & 0.481 & 0.692 & 0.545 & 0.810 & \underline{0.913} \\
    \rowcolor{white!70!yellow}
    \texttt{DLM4G}-1.o & 50M & \underline{0.534} & \underline{0.715} & \underline{0.560} & \underline{0.816} & 0.901 \\
    \rowcolor{white!70!yellow}
    \texttt{DLM4G}-2.o & 63M & \textbf{0.567} & \textbf{0.734} & \textbf{0.626} & \textbf{0.843} & \textbf{0.925} \\
    \rowcolor{white!70!gray}
    \%Gain & \textbf{x12\(\uparrow\)} & \textbf{+17.8\%} & \textbf{+6.1\%} & \textbf{+14.8\%} & \textbf{+4.1\%} & \textbf{+1.3\%} \\
    \bottomrule
    \end{tabular}%
    }
\end{table}

\noindent\textbf{Example.}
Figure~\ref{finale} shows captions produced by \texttt{DLM4G}-1.o (50M) and \texttt{DLM4G}-2.o (63M) alongside the ground truth
for a polybrominated biphenyl (PBB) molecule (SMILES shown beneath the 3D rendering). Both outputs are nearly identical and capture
(i) the molecule class (PBBs as biphenyl derivatives), (ii) the substitution range (1--10 bromine atoms), and (iii) the typical use case
(flame-retardant additives in plastics).

\noindent Quantitatively, Table~\ref{tab:single_dataset_comparison} shows that the two variants are close across BLEU, chrF++, and METEOR,
indicating that the smaller 50M model is already strong on this task, while \texttt{DLM4G}-2.o attains the best overall performance.
Additional dataset statistics and implementation details are provided in the main paper.





\end{document}